\documentclass[lettersize,journal]{IEEEtran}

\usepackage{amsmath,amsfonts}
\usepackage{algorithmic}
\usepackage{array}
\usepackage{textcomp}
\usepackage{stfloats}
\usepackage{url}
\usepackage{verbatim}
\usepackage{graphicx}
\usepackage{cite}
\usepackage{xcolor}
\usepackage{caption}
\usepackage{multirow}
\usepackage{booktabs}
\usepackage{soul,xcolor}
\usepackage[ruled]{algorithm2e}
\usepackage[colorlinks,
            linkcolor=blue,
            anchorcolor=blue,
            citecolor=red]{hyperref}
\usepackage{subfigure}
 \usepackage{array}
            
\begin{document}
\setstcolor{blue}

\title{Tight Fusion of Events and Inertial Measurements for Direct Velocity Estimation}

\author{Wanting Xu,~\IEEEmembership{Student Member,~IEEE,}
Xin Peng,~\IEEEmembership{Student Member,~IEEE,}
and Laurent Kneip,~\IEEEmembership{Senior Member,~IEEE}
\thanks{}
\thanks{W.~Xu and X.~Peng are with the Mobile Perception Lab of School of Information Science and Technology, ShanghaiTech University, Shanghai 201210, China; X.~Peng is also with Inc. Motovis, Shanghai, China.

(e-mail:~xuwt@shanghaitech.edu.cn; pengxin1@shanghaitech.edu.cn).

L.~Kneip is the corresponding author and with the Mobile Perception Lab of the School of Information Science and Technology, ShanghaiTech University, Zhong Ke Rd 1 (SIST 1C-303E), Shanghai 201210, China. (e-mail:~lkneip@shanghaitech.edu.cn).
}}



\maketitle
\begin{abstract}
Traditional visual-inertial state estimation targets absolute camera poses and spatial landmark locations while first-order kinematics are typically resolved as an implicitly estimated sub-state. However, this poses a risk in velocity-based control scenarios, as the quality of the estimation of kinematics depends on the stability of absolute camera and landmark coordinates estimation. To address this issue, we propose a novel solution to tight visual-inertial fusion directly at the level of first-order kinematics by employing a dynamic vision sensor instead of a normal camera. More specifically, we leverage trifocal tensor geometry to establish an incidence relation that directly depends on events and camera velocity, and demonstrate how velocity estimates in highly dynamic situations can be obtained over short time intervals. Noise and outliers are dealt with using a nested two-layer RANSAC scheme. Additionally, smooth velocity signals are obtained from a tight fusion with pre-integrated inertial signals using a sliding window optimizer. Experiments on both simulated and real data demonstrate that the proposed tight event-inertial fusion leads to continuous and reliable velocity estimation in highly dynamic scenarios independently of absolute coordinates. Furthermore, in extreme cases, it achieves more stable and more accurate estimation of kinematics than traditional, point-position-based visual-inertial odometry.
\end{abstract}


\section{Introduction}

\IEEEPARstart{O}{ver} the past two decades, monocular visual-inertial ego-motion estimation has turned from a scientific challenge into a mature, indispensable solution in restricted energy, payload, and budget applications such as low-end consumer service robotics, Unmanned Aerial Vehicles (UAVs), or intelligence augmentation devices. While popular open-source solutions mostly rely on sparse feature extraction (e.g. ORB-SLAM~\cite{mur2015orb}, VINS-Mono~\cite{qin2018vins}, OKVIS~\cite{leutenegger2015keyframe}), the community has also explored alternative feature-based methods (e.g. PL-VIO~\cite{he2018pl}, PL-SLAM \cite{pumarola2017pl}), dense optical flow-based approaches (e.g. VOLDOR+SLAM~\cite{min21voldor}), or direct photometric error minimizers (e.g. DM-VIO~\cite{stumberg22dmvio}).

In the following, the discussion distinguishes between absolute world-centric and relative camera-centric parameters. The former are given by the absolute sensor pose as well as the absolute landmark coordinates in the world (i.e. in a world reference frame), while the latter are given by the dynamic ego-state (i.e. translational and rotational velocity and acceleration expressed in the camera frame) as well as the relative depth of the landmarks with respect to the current frame. The above mentioned SLAM frameworks all employ absolute world-centric representations, and the kinematics of the sensor are estimated as implicitly estimated sub-states rather than directly measured. This enables map construction and memorization as well as global localization and path planning. The down-side of such representations, however, is that the stability of the dynamic states depends on the stability of the overall state estimation, which includes the aforementioned absolute world-centric parameters. Events such as map tracking failures or loop closures may easily impact on the quality of the dynamics estimation, and hence prevent the reliable solution of safety-critical velocity-based control problems (e.g. UAV stabilization, obstacle avoidance). This motivates fail-safe approaches in which continuous epipolar geometry and inertial cues are coupled to directly estimate the dynamics of a platform~\cite{weiss20134dof}.

\begin{figure}[t]
\centering
	\includegraphics[width=0.6\linewidth]{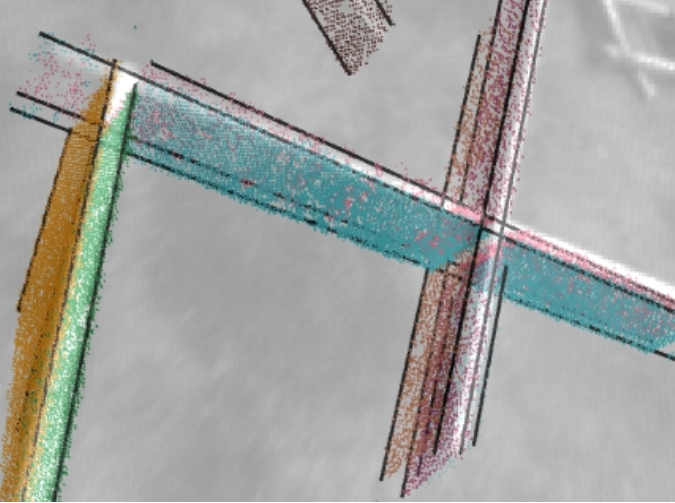}
	\caption{Example of a local event stream pattern used for camera velocity estimation. Each cluster of events corresponds to a locally observed, real-world line segment.}
\label{fig:real}
\vspace{-0.2in}
\end{figure}

In this paper, we present a novel solution to visual-inertial fusion for direct dynamics estimation by employing a dynamic vision sensor. The latter---also called an event camera---is a bio-inspired, low-latency sensor that operates fundamentally differently from a regular camera. Rather than measuring intensity images frame-by-frame, an event camera employs independent pixel-level CMOS circuits measuring the change of brightness patterns. Each individual pixel asynchronously fires time-stamped events whenever the logarithmic intensity change since the last event exceeds a predefined threshold. Event cameras have latencies in the order of micro-seconds, may fire events at a very high rate, and own high dynamic range. These properties make the event camera an excellent choice in challenging visual conditions caused by high dynamics or low illumination~\cite{zhou2021event,mueggler2018continuous,zuo2022devo}.

We consider event cameras to be a highly intuitive choice for tight visual-inertial dynamics estimation, as events are a direct consequence of velocities and relative feature depths. Our solution relies on our previous result, a novel trifocal tensor-based~\cite{hartley2003multiple} incidence relation that can be solved in closed-form and directly relates events triggered by moving appearance edges, the corresponding lines, and camera velocities~\cite{peng2021continuous}. Robustness is achieved through a novel two-layer RANSAC scheme, and a regularized, sliding-window optimizer ensures smooth velocity estimation over time. As demonstrated by our results on both simulated and real data, our approach generates velocity samples from temporal intervals of events that are comparable to the exposure time of regular images. Event camera-based visual-inertial fusion for direct velocity sensing achieves highly reliable results, and---owing to the absence of motion blur and a positive correlation between SnR and motion dynamics---appears as a highly sensible, fail-safe sensor-fusion strategy able to cope with challenging high-velocity UAV motion.

Our main contributions are listed as follows:
\begin{itemize}
    \item To the best of our knowledge, the first comprehensive work on turning the highly bio-inspired setup of an event camera and an inertial sensor into a direct line-based visual-inertial 3D speed sensor.
    \item We pioneer the use of RANSAC with events, and propose a novel nested two-layer RANSAC scheme for geometric, event-based velocity initialization from moving line observations. The outer layer is a minimal application within RANSAC of the trifocal tensor constraint inspired by our previous work~\cite{peng2021continuous}, and a novel 4-event solver for 3D line reconstruction is proposed in the inner layer.
    \item We further introduce a complete inertial event-based solution including a bootstrapper and a purely relative sliding window back-end optimizer. The regularization is formulated using manifold-based pre-integration of inertial readings.
\end{itemize}

The paper is organized as follows. Section~\ref{sec:related_work} reviews related work including real-time visual-inertial motion and structure estimation, visual-inertial speed-sensing, line-based visual-inertial fusion, and event-based visual odometry and SLAM. Section~\ref{sec:preliminaries} gives a brief review of the continuous event-line constraint and the line representation used in this work. The theory then is divided into two parts---initialization through closed-form velocity calculation in Section~\ref{sec:initialization}, and tightly-coupled back-end optimization in Section~\ref{sec:back_end_opt}. Experiments are in Sections~\ref{sec:experiment1} and \ref{sec:experiment2}.

\section{Related Work}
\label{sec:related_work}

We provide an overview of the most important state-of-the-art contributions on visual-inertial odometry and SLAM with standard cameras as well as direct visual-inertial speed estimation and line-based visual odometry and SLAM. Note that the field is very broad, and that the present overview mostly introduces recent, state-of-the-art contributions employing traditional geometric concepts. For a more complete overview including modern data-driven methods, the reader is kindly referred to~\cite{pastpresent16}. The section concludes with a review on more recent contributions on event-based motion estimation.

\subsection{Visual-inertial odometry and SLAM}

The state-of-the-art solution to monocular real-time motion and structure estimation consists of fusing the visual measurements with an inertial measurement unit (IMU). The latter is composed of an accelerometer and a gyroscope to measure body accelerations and angular velocities. IMUs provide highly complementary information to visual readings, and thus have contributed substantially to the current robustness of visual-inertial localization and mapping~\cite{pastpresent16}. Most importantly, IMUs add metric scale to the otherwise scale invariant results from monocular SLAM. The preferred fusion strategy is tightly-coupled, and was initially demonstrated in seminal works by Sterlow and Singh~\cite{sterlow04}, Dong-Si and Mourikis~\cite{dongsi11}, and Mourikis et al.~\cite{mourikis2007multi}. The first two works propose the first optimization-based visual-inertial fusion techniques (the latter one including fixed-lag smoothing). The third work proposes MSCKF, a popular EKF-based visual-inertial odometry system which updates IMU error states alongside camera poses. Based on the findings of Strasdat et al.~\cite{strasdat10}, optimization-based methods have become the go-to strategy for visual-inertial fusion. Furthermore, the works of Lupton and Sukkarieh~\cite{lupton2011visual} and Forster et al.~\cite{forster16} have introduced IMU pre-integration terms, thus avoiding repeated integrations during optimization. These findings have nowadays led to the state-of-the-art optimization-based visual-inertial fusion frameworks ORB-SLAM3~\cite{campos2021orb}, VINS-Mono~\cite{qin2018vins}, and OKVIS~\cite{leutenegger2015keyframe}. More recently, the community has also introduced dense optical flow~\cite{min21voldor} and direct, photometric fusion techniques~\cite{stumberg22dmvio}. The above-mentioned frameworks all employ world-centric coordinates, and as such the stability of the estimation is highly dependent on reliable local map tracking. Furthermore, the use of normal cameras and their tendency of producing blurry images naturally puts limitations to the tolerable motion dynamics.

\subsection{Direct visual-inertial velocity estimation}

The ego-velocity can be obtained as an implicitly estimated sub-state estimated in position-based visual-inertial fusion, by conducting temporal differentiation of positions, or by direct estimation from sensor measurements. Most velocity estimation algorithms rely on the relationship between pixel velocities (optical flow) and metric velocities~\cite{song2007vision,honegger2012real,honegger2013open,weiss2012real,weiss20134dof}. PX4FLOW~\cite{honegger2013open} is a popular optical flow sensor that consists of a camera, a gyroscope and an ultrasonic range sensor. The velocity estimation of PX4FLOW requires stable depth readings and highly depends on the planarity of the observed scene. Song et. al.~\cite{song2007vision} compute the 2D velocity of a downward-facing camera from optical flow measurements using a known scene depth assumption. Weiss et. al.~\cite{weiss2012real} propose an inertial-optical flow framework for metric 3D speed estimation of a self-calibrating camera-IMU setup. The camera is regarded as a speed sensor, and the algorithm makes use of the continuous 8-point algorithm~\cite{ma2004invitation} for scale-invariant velocity samples. The combined use of optical flow and IMU measurements achieves complete dynamic vehicle state estimation~\cite{weiss20134dof}. More recently, Deng et. al.~\cite{deng2018visual} propose a multicopter metric velocity estimation algorithm which also combines a low-cost IMU and a monocular camera. Outliers in the point correspondences are removed by the Mean Shift algorithm, and the internal estimator is given by a Linear Kalman Filter. Moreover, Gao et al.~\cite{gao2019efficient} rely on optical flow extracted from a forward-looking stereo camera to estimate the translational velocity of as MAV. Different from the aforementioned velocity estimation approaches, our method estimates speed by fusing an event camera and an IMU, thus achieving better performance in challenging dynamic scenarios.

\subsection{Line-based multiple-view geometry and visual SLAM}

Besides an abundance of works on sparse point-based structure-from-motion, visual odometry and SLAM, the community has invested significant efforts into the development of higher-level feature-based implementations. Most commonly, frameworks include lines and planes to represent larger segments of the environment. They are registered against straight line measurements or uniformly colored image segments such as super-pixels, and generally increase the algorithm’s robustness and accuracy in otherwise feature-deprived scenarios. Weng et al.~\cite{weng1992motion} propose a closed-form solution for pose estimation with line correspondences. As demonstrated, at least three views are needed to do projective reconstruction from line correspondences. Hartley further discusses the trifocal tensor, an algebraic object that helps to link the motion in three views to point or line-feature observations~\cite{hartley1997lines}. All details about trifocal tensor geometry with lines can be found in the book by Hartley and Zisserman~\cite{hartley2003multiple}. Line detection and description can be done using the state-of-the-art LSD line detector~\cite{von2008lsd}. Typically, high-level features are utilized in conjunction with points. PL-SLAM~\cite{pumarola2017pl} is proposed to handle low-texture scenes by merging line features into ORB-SLAM. Tightly-coupled monocular visual–inertial odometry with points and lines is further proposed by He~et~al.~\cite{he2018pl}, who minimize both IMU integration errors and reprojection errors over points and lines.

Event cameras generate events for changing brightness levels at every pixel. As such, events are mostly generated by moving high-gradient regions in the image, i.e. they are mostly sensitive to moving appearance edges. Given the abundance of straight lines in man-made environments and the compactness of line representations, we utilize them in our event-inertial fusion framework to model event-generating segments of the environment in 3D.

\subsection{Event-based motion estimation}

A thorough investigation of event-based vision is provided in the survey of Gallego et al.~\cite{gallego2020event} or via online resources~\cite{github:event_resources}. Event cameras became more popular about a decade ago. However, the fundamentally different, asynchronous nature of event streams makes it difficult to directly use traditional geometric constraints from multiple-view geometry. Instead of full 6-Dof SLAM systems, most original works on event-based motion estimation therefore focus on simpler scenarios. Weikersdorfer et. al.~\cite{weikersdorfer2013simultaneous} originally propose a 2D-SLAM system with a dynamic vision sensor by employing a particle filter. The same group also proposes an event-based 3D SLAM framework by fusing events with a classical frame-based RGB-D camera~\cite{weikersdorfer2014event}. Other event-based visual odometry systems make use of known depth or 3D structure~\cite{censi2014low,mueggler2014event,gallego2016event,chamorro2020high,bryner2019event}, or are simply limited to the pure rotation scenario~\cite{gallego2018unifying}. Contrast maximization~\cite{gallego2018unifying,gallego2019focus} is proposed as a unifying framework applicable to several event-based vision tasks. Although it has garnered significant attention from researchers in the field~\cite{kueng2016low,liu2020globally,peng2021globally}, its applicability is currently restricted to homographic warping scenarios. Full 6-DoF estimation is solved by Kim et al.~\cite{kim2016real} using a filtering approach, and Rebecq et al.~\cite{rebecq2016evo} use an alternating tracking and mapping framework. Zhu et al.~\cite{zhu17} and Rebecq et al.~\cite{rebecq2017real} furthermore propose more reliable frameworks by fusing the measurements with an IMU. Furthermore, Mueggler et. al.~\cite{mueggler2018continuous} leverage continuous-time representations and spline-based trajectory optimization to perform visual-inertial odometry with an event camera.

More reliable 6-DoF odometry and SLAM solutions keep being obtained by fusion with other sensors. Kueng et al.~\cite{kueng2016low} combine the event camera with a standard camera to track features and build a probabilistic map. A similar sensor combination is used in Ultimate-SLAM~\cite{vidal2018ultimate}, which improves robustness and accuracy by a combined minimization of both vision and event-based residual errors. Zhou et al.~\cite{zhou2021event} propose the first event-based stereo odometry system. Zuo et al.~\cite{zuo2022devo} propose the use of a hybrid stereo setup of an event and a depth camera to realize DEVO, a semi-dense edge-tracking method inspired by Canny-VO~\cite{zhou2018canny}, and Hidalgo-Carrió et al.~\cite{hidalgo2022event} introduce EDS, a 6-DOF monocular direct visual odometry which combines events and frames.

Event-based motion estimation can be divided into optimization-based~\cite{rebecq2016evo,le2020idol,mueggler2018continuous}, filter-based~\cite{zihao2017event,weikersdorfer2013simultaneous} and learning-based~\cite{maqueda2018event,gehrig2020event} solutions. However, there is a lack of research on how fundamental geometry can be applied to event-based vision. In this work, we leverage our previous result~\cite{peng2021continuous} on applying trifocal tensor geometry~\cite{hartley2003multiple} to explain the relationship between the events generated by a 3D line feature observation and the ego-motion of an event camera. We extend our previous result by a more reliable direct solution of the camera velocity, and furthermore propose a novel visual-inertial fusion back-end to achieve reliable, velocity estimation. The back-end is similar to the line-based visual-inertial odometry framework IDOL~\cite{le2020idol}, except that we directly perform dynamics estimation in camera-centric coordinates, a more intuitive and fail-safe approach that does not depend on stable global map tracking.

\section{Preliminaries}
\label{sec:preliminaries}

We start by reviewing the Continuous Event-Line Constraint (CELC), a fundamental geometric incidence relationship that establishes a link between events and the translational velocity of the camera. Given the difficulty of extracting and matching sparse features in event streams and the fact that events present high sensitivity to edges, we choose a higher-level representation of the environment in order to formulate the geometry: lines. The section therefore also briefly reviews two representations of 3D lines---Pl\"ucker line coordinates and their orthonormal representation---which are utilized within the later nonlinear optimization back-end. The notation used throughout the paper is introduced alongside the theory.

\subsection{The Continuous Event-Line Constraint (CELC)}
\label{sec:eventline_celc}

CELC is a constraint proposed in our previous work \cite{peng2021continuous}. It expresses the relation between the events generated by a moving observed line and the event camera's ego-motion using trifocal tensor geometry. Rather than using individual line feature detections as for example given by the ELiSeD detector by Brandli et al.~\cite{brandli2016elised}, we adopt the more general method of Le Gentil et al.~\cite{le2020idol}, which identifies clusters in the space-time volume of events, each one being generated by a moving line observation in the image (i.e. a moving straight appearance boundary). By using this method, the event data is essentially left in its original form.

\begin{figure}[bt]
\centering
	\includegraphics[width=0.8\linewidth]{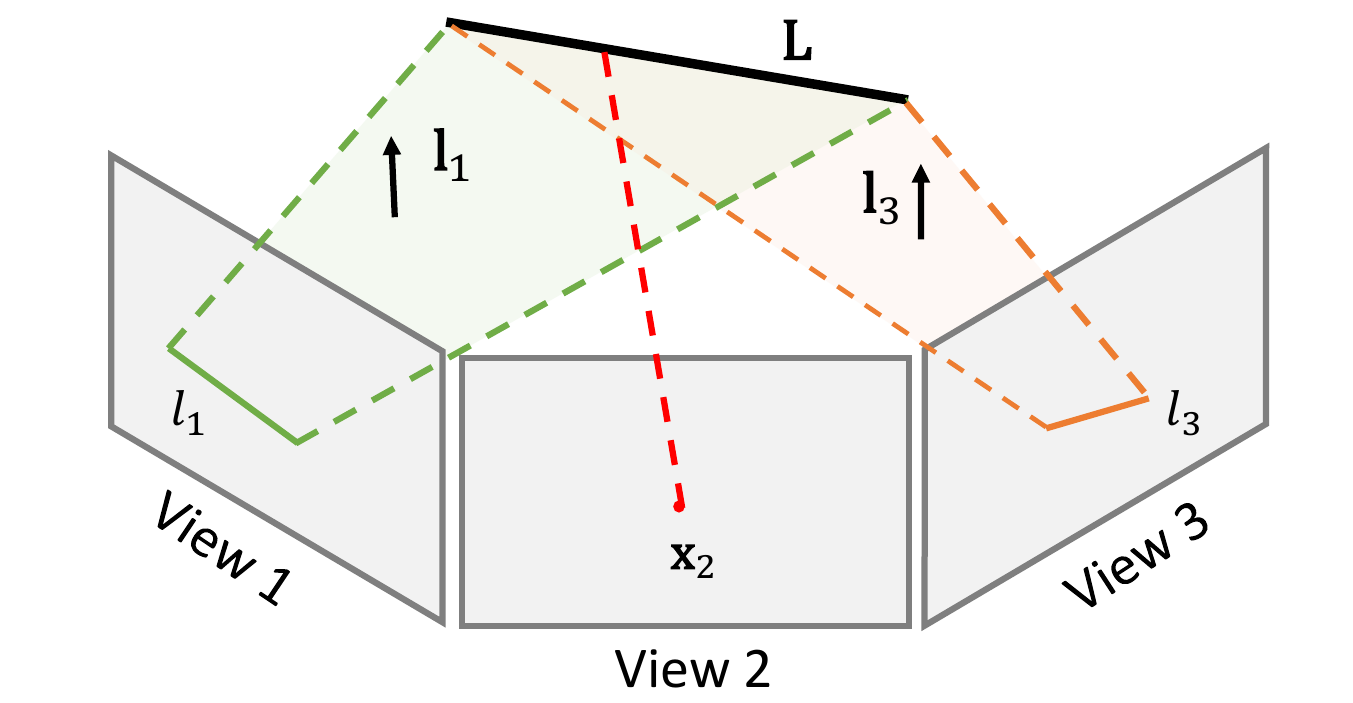}
	\includegraphics[width=0.8\linewidth]{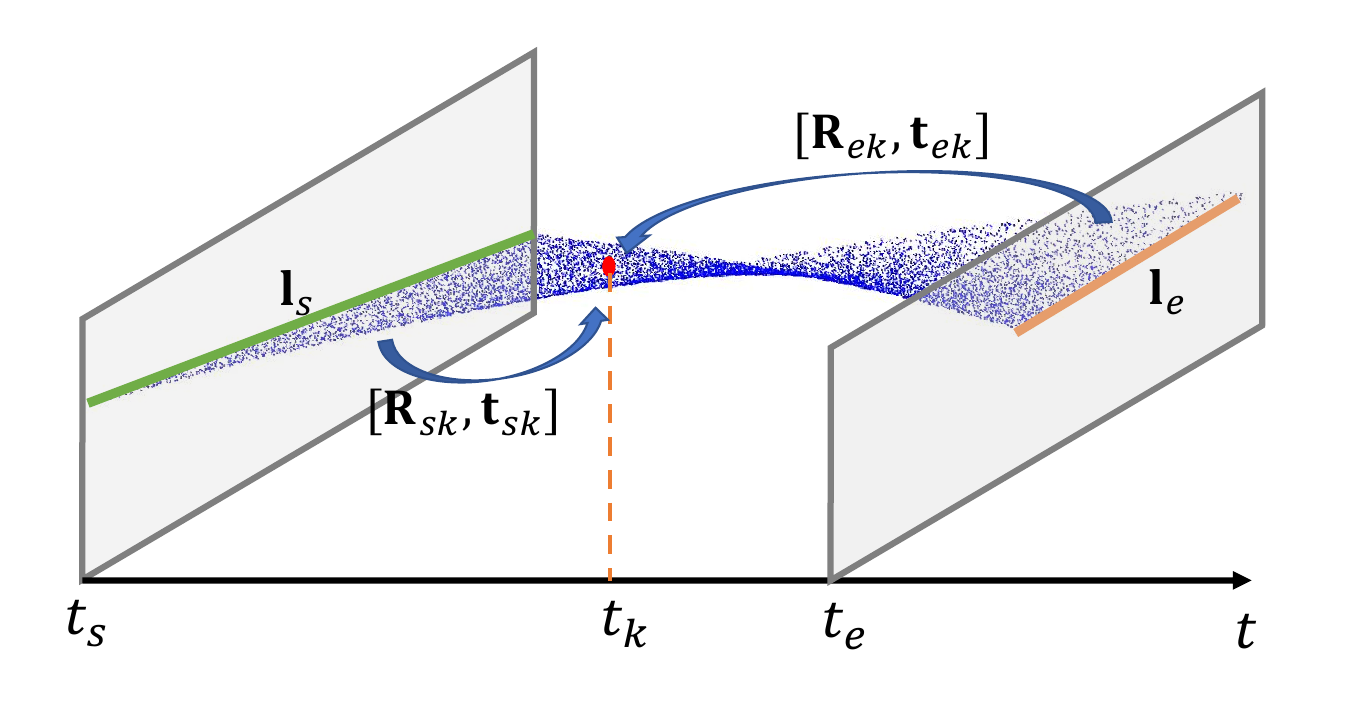}
	\caption{Geometry of CELC which indicates the relationship between events, lines and camera ego-motion. In our work, we extract the intermediate line representations $\mathbf{l}_{sj}$ and $\mathbf{l}_{ej}$ by choosing two events in a small time interval $\Delta t$ at the beginning and at the end of the interval.}
\label{fig:CELC}
\end{figure}

Let us denote the $j$-th cluster of events $\mathcal{E}_{j}$. Let $\mathcal{E}_{j}$ span a time interval $[t_{sj},t_{ej}]$ in the space-time volume of events, and let $e_{ij}$ furthermore represent the $i$-th event in the cluster. A 3D line has 4 degrees of freedom, which we parametrize by the projections of the line onto the virtual camera frames at timestamps $t_{sj}$ and $t_{ej}$. Let $\mathbf{l}_{sj}=\mathbf{K}^T \boldsymbol{\lambda}_{sj}$ and $\mathbf{l}_{ej}=\mathbf{K}^T \boldsymbol{\lambda}_{ej}$ denote the normalized plane coordinates corresponding to these line observations, where $\boldsymbol{\lambda}_{sj}$ and $\boldsymbol{\lambda}_{ej}$ denote the image-based representation of those lines (note that in this work $\boldsymbol{\lambda}_{sj}$ and $\boldsymbol{\lambda}_{ej}$ are only auxiliary variables representing the unknown line in 3D, and they will later on be hypothesized as a function of individual events), and $\mathbf{K}$ denotes an upper-triangular intrinsic camera matrix.

Finally, let us assume that the camera exhibits constant velocity motion in the local camera frame during the entire time interval. For any event $e_{ij}=\{x_{ij},y_{ij},t_{ij},s_{ij}\}$ with timestamp $t_{ij}$, normalized coordinates $\mathbf{f}_{ij}=\mathbf{K}^{-1}[x_{ij}\text{ }y_{ij}\text{ }1]^T$, and polarity $s_{ij}$ (unused in the algorithm, but used when tracking lines.), we then have
\begin{eqnarray}
\mathbf{f}_{ij}^{\mathsf{T}} \mathbf{B}_{ij} \mathbf{v} = 0,
\label{eqn:EventRANSAC_CELC}
\end{eqnarray}
where 
\begin{eqnarray}
 \mathbf{B}_{ij} = \left[ \begin{matrix}
    (t_{ij} - t_{ej}) \mathbf{l}_{sj}^\mathsf{T} \mathbf{r}_1^{sij} \mathbf{l}_{ej}^\mathsf{T}   - (t_{ij} - t_{sj}) \mathbf{l}_{ej}^{\mathsf{T}} \mathbf{r}_1^{eij} \mathbf{l}_{sj}^\mathsf{T}  \\
    (t_{ij} - t_{ej}) \mathbf{l}_{sj}^\mathsf{T} \mathbf{r}_2^{sij} \mathbf{l}_{ej}^\mathsf{T}  - (t_{ij} - t_{sj}) \mathbf{l}_{ej}^{\mathsf{T}} \mathbf{r}_2^{eij} \mathbf{l}_{sj}^\mathsf{T}  \\
    (t_{ij} - t_{ej}) \mathbf{l}_{sj}^\mathsf{T} \mathbf{r}_3^{sij} \mathbf{l}_{ej}^\mathsf{T} - (t_{ij} - t_{sj}) \mathbf{l}_{ej}^{\mathsf{T}} \mathbf{r}_{3}^{eij} \mathbf{l}_{sj}^\mathsf{T} \\
    \end{matrix}
    \right].
\end{eqnarray}
$\mathbf{r}_{1}^{sij},\, \mathbf{r}_{2}^{sij},\, \mathbf{r}_{3}^{sij}$ mark the columns of the rotation $\mathbf{R}_{sij} = \exp(\lfloor\boldsymbol{\omega}\rfloor_{\times}(t_{ij} - t_{sj}))$ from the camera pose at time  $t_{ij}$ to the pose at time $t_{sj}$, while $\mathbf{r}_{1}^{eij},\, \mathbf{r}_{2}^{eij},\, \mathbf{r}_{3}^{eij}$ are the columns of the rotation $\mathbf{R}_{eij} = \exp(\lfloor\boldsymbol{\omega}\rfloor_{\times}(t_{ij} - t_{ej}))$ from the camera pose at time  $t_{ij}$ to the pose at time $t_{ej}$. Here we use $\boldsymbol{\omega}$ to represent the angular velocity, and $\lfloor\boldsymbol{\omega}\rfloor_{\times}$ is the $3\times3$ matrix skew symmetric matrix form of $\omega$. $\mathbf{t}_{sij} = (t_{ij} - t_{sj})\mathbf{v}$ and $\mathbf{t}_{eij} = (t_{ij} - t_{ej})\mathbf{v}$ are the corresponding translations of these relative poses. $\boldsymbol{\omega}$ denotes the known rotational velocity taken from the IMU, and $\mathbf{v}$ the translational velocity that we wish to identify. The constraint is derived from a continuous-time adaptation of the trifocal tensor, which uses a constant-velocity motion assumption. The geometry is illustrated in Fig.~\ref{fig:CELC}. 

Note that in the following, we assume that the camera and the IMU are mounted close enough such that the physical distance between the two sensors can be ignored and set to 0. Note furthermore that we assume that the system is calibrated and that the rotational transformation between the IMU and the camera is known. Without loss of generality, IMU signals are assumed to be pre-rotated and the camera and IMU frames are assumed to be identical (henceforth denoted as the body frame).

Given a sequence of events with $M$ line clusters $\mathcal{E}_j$ where $j=1,2...,M$, the constraints from each event cluster with $N_{j}$ events can be stacked into the single linear problem 
\begin{equation}
    \left[ \begin{matrix}
    \mathbf{B}_{11}^{\mathsf{T}} \mathbf{f}_{11}~
    \dots ~
    \mathbf{B}_{ij}^{\mathsf{T}} \mathbf{f}_{ij}  ~
    \dots ~
    \mathbf{B}_{N_M M}^{\mathsf{T}} \mathbf{f}_{N_M M}
    \end{matrix} \right]^\mathsf{T} \mathbf{v} = \mathbf{0}.
    \label{eqn:linear_constraint}
\end{equation}
With known angular velocity, (\ref{eqn:linear_constraint}) is linear in $\mathbf{v}$. By ignoring the presence of outliers and assuming known line reprojections at $t_{sj}$ and $t_{ej}$, the translational velocity under an algebraic error criterion can be obtained from a simple SVD. Note that, in analogy to monocular structure-from-motion, results are determined only up to an unknown scale factor.

\subsection{Line Representation Methods}
\label{sec:line_rep}

A 3D line has 4 degrees of freedom (DoF). 3D lines may be represented in various ways~\cite{bartoli2005structure}. One of the most common representations is given by Pl\"ucker line coordinates, a representation that easily enables geometric 3D lines transformations as linear operations. However, the representation is non-minimal and has 6 DoF. In order to perform minimal 4-DoF updates within non-linear optimization---a key requirement in back-end optimization---we therefore also make use of the orthonormal representation proposed by Bartoli et al.~\cite{bartoli2005structure}. The combined use of both representations is consistent with the works of Zhang et al.~\cite{zhang2015building} and He et al.~\cite{he2018pl}. In the following, we will review both of them.

\subsubsection{Pl\"ucker line coordinates}

\begin{figure}[t]
    \centering
    \includegraphics[width=0.7\linewidth]{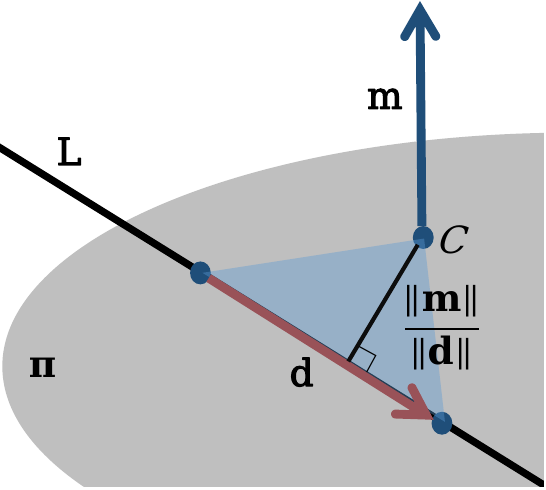}
    \caption{The geometry of Pl\"ucker line coordinates.}
    \label{fig:plucker}
\end{figure}

As shown in Fig.~\ref{fig:plucker}, a 3D line $\mathbf{L}$ is represented by the Pl\"ucker line coordinates $\mathbf{L} = [\mathbf{d}^\top, \mathbf{m}^\top]^\top \in \mathbb{R}^6$, where the 3D vector $\mathbf{d}$ is the direction vector of the line, and the 3D vector $\mathbf{m}$ is the moment vector which is normal to the plane $\boldsymbol{\pi}$ determined by line $\mathbf{L}$ and the origin. $\mathbf{m}$ is perpendicular to $\mathbf{d}$ ($\mathbf{m}^\top \mathbf{d} = 0$), and is defined as
\begin{equation}
    \mathbf{p}\times\mathbf{d} = \mathbf{m},
    \label{equ:md}
\end{equation}
where $\mathbf{p}$ is a vector from $C$ to any arbitrary point on line $\mathbf{L}$. Note that $\mathbf{m}$ and $\mathbf{d}$ do not need to be unit vectors, but their ratio $\|\mathbf{m}\| / \|\mathbf{d}\|$ defines the orthogonal distance between the origin $C$ and the line $\mathbf{L}$. We typically set $\mathbf{d}$ as a unit vector, that is $\|\mathbf{d}\| = 1$. Naturally, the 6-dimensional Pl\"ucker vector has two side constraints, which is in agreement with the 4 DoF of a 3D line.

While the existence of side constraints make Pl\"ucker line coordinates inconvenient to be applied in nonlinear optimization, they are very useful to construct geometric transformations with lines. For a given transformation from frame $i$ to frame $j$, with rotation $\mathbf{R}_{ji}$ and translation $\mathbf{t}_{ji}$, the corresponding line geometry transformation~\cite{bartoli20013d} using Pl\"ucker line coordinates is
\begin{equation}
\label{equ:linetrans}
    \begin{bmatrix}
    \mathbf{m}_j \\
    \mathbf{d}_j 
    \end{bmatrix}
    =
    \begin{bmatrix}
    \mathbf{R}_{ji} & [\mathbf{t}_{ji}]_{\times}\mathbf{R}_{ji} \\
    \mathbf{0} & \mathbf{R}_{ji} 
    \end{bmatrix}
    \begin{bmatrix}
    \mathbf{m}_i \\
    \mathbf{d}_i 
    \end{bmatrix}.
\end{equation}

\subsubsection{Orthonormal representation}

Based on Pl\"ucker line coordinates, the orthonormal representation is a minimal representation of a 3D line that can be regarded as the tangential space around a given Pl\"ucker coordinate. It is well suited for nonlinear optimization, and can be converted back-and-forth to regular Pl\"ucker line coordinates. For a given Pl\"ucker line $\mathbf{L} = [\mathbf{d}^\top, \mathbf{m}^\top]^\top$, the corresponding orthonormal representation $(\mathbf{U}, \mathbf{W}) \in SO(3) \times SO(2)$ can be obtained by using the QR decomposition 

\begin{equation}
    \begin{bmatrix}
    \mathbf{m} &
    \mathbf{d} 
    \end{bmatrix}_{3\times 2}
    = \mathbf{U}_{3\times 3} \mathbf{\Sigma}_{3\times 2},
\end{equation}
where 
\begin{eqnarray}
    \mathbf{U}_{3\times 3}
    &=&
    \begin{bmatrix}
    \frac{\mathbf{m}}{\parallel \mathbf{m} \parallel} & \frac{\mathbf{d}}{\parallel \mathbf{d} \parallel} & \frac{\mathbf{m} \times \mathbf{d} }{\parallel \mathbf{m} \times \mathbf{d} \parallel}
    \end{bmatrix}, \nonumber\\
    \mathbf{\Sigma}_{3\times 2} &=&  \begin{bmatrix}
    \parallel \mathbf{m} \parallel  & 0\\
    0 & \parallel \mathbf{d} \parallel\\
    0 & 0
    \end{bmatrix}.
\end{eqnarray}

Note that vector $(\parallel \mathbf{m} \parallel, \parallel \mathbf{d} \parallel)^\mathsf{T}$ has only one DoF~\cite{bartoli2005structure,he2018pl}, hence matrix $\Sigma$ can be represented in a compressed way by using an $SO(2)$ matrix
\begin{eqnarray}
    \mathbf{W} &=& \frac{1}{\sqrt{ \parallel \mathbf{m} \parallel^2+ \parallel \mathbf{d} \parallel^2}}
    \begin{bmatrix}
         \parallel \mathbf{m} \parallel & - \parallel \mathbf{d} \parallel \\
          \parallel \mathbf{d} \parallel &  \parallel \mathbf{m} \parallel
    \end{bmatrix} \nonumber\\
    &=& \begin{bmatrix}
         \cos(\phi) & - \sin(\phi) \\
        \sin(\phi) &   \cos(\phi)
    \end{bmatrix}
        =
    \begin{bmatrix}
    w_1 & -w_2\\
    w_2 & w_1
    \end{bmatrix},
\end{eqnarray}
where $\phi$ is a rotation angle. $\mathbf{U}$ and $\mathbf{W}$ are actually rotation matrices of three and two dimensions respectively, and they can be locally updated during optimization using

\begin{equation}
  \begin{gathered}
    \mathbf{U} \leftarrow \mathbf{U}\mathbf{R}(\boldsymbol{\theta}), \\
    \mathbf{W} \leftarrow \mathbf{W}\mathbf{R}(\theta),
  \end{gathered}
\end{equation}
with

\begin{equation}
  \begin{gathered}
    \boldsymbol{\theta} = 
    \begin{bmatrix}
    \theta_1 & \theta_2 & \theta_3
    \end{bmatrix}^\top, \\
    \mathbf{R}(\boldsymbol{\theta}) = \mathbf{R}_x(\theta_1)\mathbf{R}_y(\theta_2)\mathbf{R}_z(\theta_3),
  \end{gathered}
\end{equation}
where $\mathbf{R}_x(\theta_1)$, $\mathbf{R}_y(\theta_2)$ and $\mathbf{R}_z(\theta_3)$ are 3D rotation matrices in $SO(3)$ around the x, y, and z axes with angles $\theta_1$, $\theta_2$ and $\theta_3$, respectively. $\mathbf{R}(\theta)$ is a 2D rotation matrix in $SO(2)$. Thus, the four parameters for the construction of a minimal update are defined by $\mathbf{p} = [\boldsymbol{\theta}^\top, \theta]^\top$. The orthonormal representation can be directly converted to Pl\"ucker line coordinates by

\begin{equation}
\begin{split}
    \mathbf{L} 
    & = [\mathbf{d}^\top, \mathbf{m}^\top]^\top \\
    & = [\omega_2\mathbf{u}_2^\top, \omega_1\mathbf{u}_1^\top]^\top,
\end{split}
\end{equation}
where $\mathbf{u}_1$ and $\mathbf{u}_2$ are the first and second columns of $\mathbf{U}$ respectively.

\section{Initialization}
\label{sec:initialization}

We divide the whole pipeline into two parts: initialization and back-end optimization. Note that most event-based visual inertial systems pay little attention to the initialization part. For instance, IDOL~\cite{le2020idol} does not address initialization, and the sensor is assumed to be static for initialization in~\cite{vidal2018ultimate}.  We adopt the method of Le Gentil et al.~\cite{le2020idol} to cluster the events into groups each one generated by one moving line observation. We operate over a thin temporal slice (about 0.1s$\sim$0.2s) of the events, and only consider clusters that stretch over the entire slice. In the following, the $M$ clusters in discussion are simply the clusters that contain events over the temporal window. The starting and ending times of all clusters can hence be assumed equal to the starting and ending time of the temporal slice, i.e. $t_{sj} = t_s ,~ t_{ej} = t_e ,~  \forall{j=1,\ldots,M}$. We then make use of CELC (cf. Section~\ref{sec:eventline_celc}) within a two-layer nested RANSAC scheme for line-based translational velocity bootstrapping. Rather than using all events to construct Eq. (\ref{eqn:linear_constraint}), we create a hypothesis for the linear velocity $\mathbf{v}$ by sampling a small set of events from two different line clusters within the outer RANSAC layer. The hypothesised linear velocity is then verified as part of the inner RANSAC layer, which aims at robust regression of each 3D line using a novel 4-event minimal solver. The average geometric support expressed by the individual inlier ratios then implicitly makes a statement about the quality of the hypothesised dynamic motion parameters. In the following, we will introduce the detailed functionality of both layers.

\subsection{Outer Layer RANSAC}

\begin{algorithm}[t]
        \caption{Outer Layer RANSAC for Event-based Linear Velocity Estimation}
        \KwIn{Event sequence, angular velocity $\boldsymbol{\omega}$ and intrinsic matrix $K$}
        
        \KwOut{Linear velocity $\mathbf{v}$}
        Line clustering to get $M$ event clusters $\mathcal{E}_j$ where $j=1,2...,M$ \\
        \While{$p_{mean} < p_{thres}$ $\&\&$ $k$ $<$ Max Iteration}
        {
            Sample 2 lines (clusters) and 5 events per line; \\
            Find $\mathbf{v}$ by linear solver in~(\ref{eqn:linear_constraint}), $\left \|\mathbf{v}\right \| = 1$;\\
            
            \For{each cluster}
            {
                Inner layer RANSAC (Alg.~\ref{alg:inner_RANSAC}) to get $\mathbf{L}$ and $p$; \\
            }
            $k$++;\\
            Get average inlier percentage $p_{mean}$ in Sec. \ref{sec:convergence};\\
    }
    \Return{$\mathbf{v}$.}
    \label{alg:outer_RANSAC}
\end{algorithm}

We first introduce the outer layer RANSAC as described in Alg.~\ref{alg:outer_RANSAC}. Note that in order to solve (\ref{eqn:linear_constraint}), at least 2 rows are required. Note furthermore that using constraints from a single cluster is not enough, as it is intuitively clear that this would leave the component of the velocity that is parallel to that 3D line unobserved~\cite{stoffregen2019event1}. This problem is also known as the \textit{aperture problem}. We therefore need to sample events from at least two clusters. We start by randomly sampling 2 clusters $\mathcal{E}_{j_1}$ and $\mathcal{E}_{j_2}$ from the entire set of clusters, and use each one to construct one row in (\ref{eqn:linear_constraint}). In each cluster, we then randomly sample 5 events. The first two events---$e_{i_1j}$ and $e_{i_2j}$--- are located at the beginning of the time interval to form the line observation $\mathbf{l}_{sj} = \mathbf{K}^{\top}( [x_{i_1j}\text{ }y_{i_1j}\text{ }1]^{\top} \times [x_{i_2j}\text{ }y_{i_2j}\text{ }1]^{\top})$. The last two events---$e_{i_4j}$ and $e_{i_5j}$---are located at the end of the time interval and used to form the line observation $\mathbf{l}_{ej} = \mathbf{K}^{\top}( [x_{i_4j}\text{ }y_{i_4j}\text{ }1]^{\top} \times [x_{i_5j}\text{ }y_{i_5j}\text{ }1]^{\top})$. The center event is finally used to substitute $\mathbf{f}_{ij}$ by $\mathbf{f}_{i_3j}$ in (\ref{eqn:EventRANSAC_CELC}), and construct the CELC constraint. Note that, in order to make $\mathbf{l}_{sj}$ and $\mathbf{l}_{ej}$ as accurate as possible, the first two events need to be sufficiently close in time and sufficiently spaced in the image. We request a pixel distance of at least 3 pixels, and constrain the events to lie in a small sub-interval $\Delta t$ at the beginning of the interval $[t_{s},t_{e}]$ (cf. Fig.~\ref{fig:CELC}). Similar conditions are imposed on the last two events, except that the sub-interval is located towards the end of $[t_{s},t_{e}]$. To conclude, the center event is constrained to lie within $[t_{s}/3,2t_{e}/3]$. Adding these constraints improves the conditioning of the CELC constraint by avoiding near degenerate cases, such as the camera moving along a straight line with out rotation~\cite{peng2021continuous}.

\subsection{Inner Layer RANSAC}
\label{sec:lineransac}

With an initial velocity hypothesis in hand, we proceed to the inner RANSAC layer which aims at robust regression of each 3D line. The average geometric support (i.e. inlier ratio) for each line then later serves as a criterion to judge the quality of the initial velocity hypothesis. We adopt the classical RANSAC framework \cite{fischler1981random} for the 3D line estimation and propose a 4-event, closed-form minimal solver to hypothesize 3D lines. The algorithm is outlined in Alg.~\ref{alg:inner_RANSAC}. Given that the processing is individual for each cluster, index $j$ is dropped in the following.

\begin{algorithm}[t]
        \caption{Inner Layer RANSAC for 3D Line Estimation}
        \KwIn{event cluster $\mathcal{E}$, angular velocity $\boldsymbol{\omega}$, linear velocity $\mathbf{v}$ and intrinsic matrix $K$}
        
        \KwOut{Pl\"ucker line $\mathbf{L}$ and inlier ratio $p$}
         
        \While{$k$ $<$ Max Iteration}
        {
            Sample 4 events;\\
            Solve Pl\"ucker line by the event-based minimal line solver in Sec. \ref{sec:minLine} with constraints (\ref{eqn:lineconstraint}) and (\ref{eqn:2constraints}); \\
            Evaluate by the angular distance metric from Sec. \ref{sec:metricRANSAC};\\
            $k$++;
        }
    
    \Return{$\mathbf{L}$, $p$.}
    \label{alg:inner_RANSAC}
    
\end{algorithm}

\subsubsection{Event-based Minimal Line Solver}
\label{sec:minLine}
Given $\boldsymbol{\omega}$ and $\mathbf{v}$, each $\mathbf{f}_i$---the normalized coordinates of $e_i$ pointing at a 3D point on the 3D line $\mathbf{L}$ from the camera pose at $t_i$---may be compensated by the rotation $\mathbf{R}_{si}$, thus resulting in $\mathbf{f}_i^{\omega} = \mathbf{R}_{si}\mathbf{f}_i$. We may furthermore obtain the start of this spatial direction vector by utilizing the camera center at time $t_i$ given by $\mathbf{t}_{si} = \mathbf{v}(t_i-t_s)$.

\begin{figure}[b]
\vspace{-0.1in}
	\begin{center}
		\includegraphics[width=0.8\linewidth]{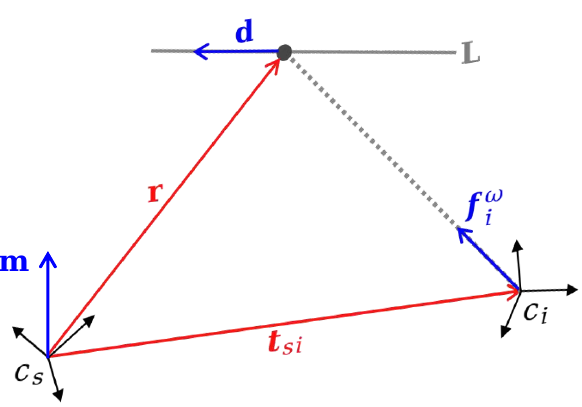}
	\end{center}
	\vspace{-0.1in}
	\caption{Geometry of the Pl\"ucker coordinates based line solver. The unknowns are the Pl\"ucker line, given by $\mathbf{d}$ and $\mathbf{m} = \mathbf{r} \times \mathbf{d}$, $\mathbf{r}$ is the position of the 3D point with respect to $c_s$. The measured or known variables are the landmark observation vector $f_i^{\omega}$ and the position of the camera center $c_i$ at timestamp $t_i$ with respect to $c_s$, given by $\mathbf{t}_{si}$. }
	\label{fig:geo}
\end{figure}

The measurement of the point on $\mathbf{L}$ can hence be expressed by the Pl\"ucker line coordinates $[\mathbf{f}_i^{\omega\top}, (\mathbf{t}_{si}\times \mathbf{f}_i^{\omega})^\top]^\top$. Let $\mathbf{L} = [\mathbf{d}^\top, \mathbf{m}^\top]^\top$ furthermore be the Pl\"ucker coordinates of the 3D line. Finally, our incidence relation is simply given by the linear, Pl\"ucker coordinate-based line-line crossing constraint
\begin{equation}
\label{eqn:lineconstraint}
    \left[ \begin{matrix}
    \mathbf{f}_i^{\omega\top}, (\mathbf{t}_{si}\times \mathbf{f}_i^{\omega})^\top
    \end{matrix} \right]
    \left[ \begin{matrix}
    \mathbf{m} \\
    \mathbf{d}
    \end{matrix} \right] = \mathbf{c}^{\mathsf{T}} \mathbf{x} = 0.
\end{equation}
The geometry is illustrated in Fig. \ref{fig:geo}. By stacking the vectors $\mathbf{c}$ for four events, a $4 \times 6$ matrix $\mathbf{C}$ is obtained. We have $\mathbf{C} = \mathbf{U}\mathbf{D}\mathbf{V}^\top$ by SVD, where the diagonal entries $d_i$ of $\mathbf{D}$ are in descending numerical order. Then the general solution is $\mathbf{x} = \lambda_{n-1}\mathbf{v}_{n-1} + \lambda_{n}\mathbf{v}_{n}$, where $\mathbf{v}_{n-1}$ and $\mathbf{v}_{n}$ are the last 2 columns of $\mathbf{V}$.

The linear problem is obviously under-constrained owing to the fact that $\mathbf{L}$ has only 4 degrees of freedom, and only 4 events have been used. Extra constrains on the solution variable are needed, which are given by
\begin{equation}
\label{eqn:2constraints}
\begin{cases}
\|\mathbf{d} \|_2 = 1 \\
\mathbf{m}^\top \mathbf{d} = 0. \\
\end{cases}
\end{equation}

In order to get $\lambda_{n-1}$ and $\lambda_{n}$, we therefore need to solve the quadratic polynomial equation system given by
\begin{equation}
\label{eqn:2constraintspoly}
\begin{cases}
\mathbf{x}(\lambda_{n-1},\lambda_{n})^\top(4:6)\cdot\mathbf{x}(\lambda_{n-1},\lambda_{n})(4:6) = 1 \\
\mathbf{x}(\lambda_{n-1},\lambda_{n})^\top(1:3)\cdot\mathbf{x}(\lambda_{n-1},\lambda_{n})(4:6) = 0, \\
\end{cases}
\end{equation}
where (j : k) represent the $j$-th to $k$-th elements of one vector. Using the Sylvester Resultant~\cite{cox2013ideals} method we can eliminate $\lambda_{n-1}$ from the two polynomials in Eq. (\ref{eqn:2constraintspoly}) and obtain a fourth-degree polynomial equation in  $\lambda_{n}$, which can be further reduced to a quadratic polynomial by substituting $\gamma = \lambda_{n}^2$. The roots of the quadratic polynomial are easily obtained in closed-form.
We finally obtain 4 solution pairs \{$\lambda_{n-1}, \lambda_{n}$\}, i.e. 4 solutions for $\mathbf{x} = [\mathbf{m}^\top, \mathbf{d}^\top]^\top$. The correct solution from the 4 is selected by the following strategy.
Intuitively, there are two lines which intersect with Pl\"ucker line $[\mathbf{f}_i^{\omega\top}, (\mathbf{t}_{si}\times \mathbf{f}_i^{\omega})^\top]^\top$. The first one is the line passing through the origin of $c_s$ and $c_i$, and the other line is the one we want to solve for --- $\mathbf{L} = [\mathbf{d}^\top, \mathbf{m}^\top]^\top$.
        
\begin{itemize}
    \item The first line can be formulate as $\mathbf{m} = [0, 0, 0]^\top$, $\mathbf{d} = \pm\mathbf{v} / \mathrm{norm}(\mathbf{v})$. Theoretically, $\pm\mathbf{v}\cdot(\mathbf{t}_{si}\times \mathbf{f}^\omega_{i}) = \mathbf{f}^\omega_{i}\cdot (\pm\mathbf{v}\times \mathbf{v}(t_i-t_s)) = 0$, which makes the line meet constraint (\ref{eqn:lineconstraint}). It is easy to see that the line also fulfills constraint (\ref{eqn:2constraints}). These two solutions are easily excluded by judging if $\mathbf{d}\times \mathbf{v} = 0$.
    
    \item The remaining candidates are given by $[\mathbf{d}^\top, \mathbf{m}^\top]^\top$ and $[-\mathbf{d}^\top, -\mathbf{m}^\top]^\top$. They actually represent the same line, and we only need to randomly pick one of them.
\end{itemize}

\subsubsection{Inlier metric}
\label{sec:metricRANSAC}

\begin{figure}[b]
\vspace{-0.1in}
	\begin{center}
		\includegraphics[width=0.8\linewidth]{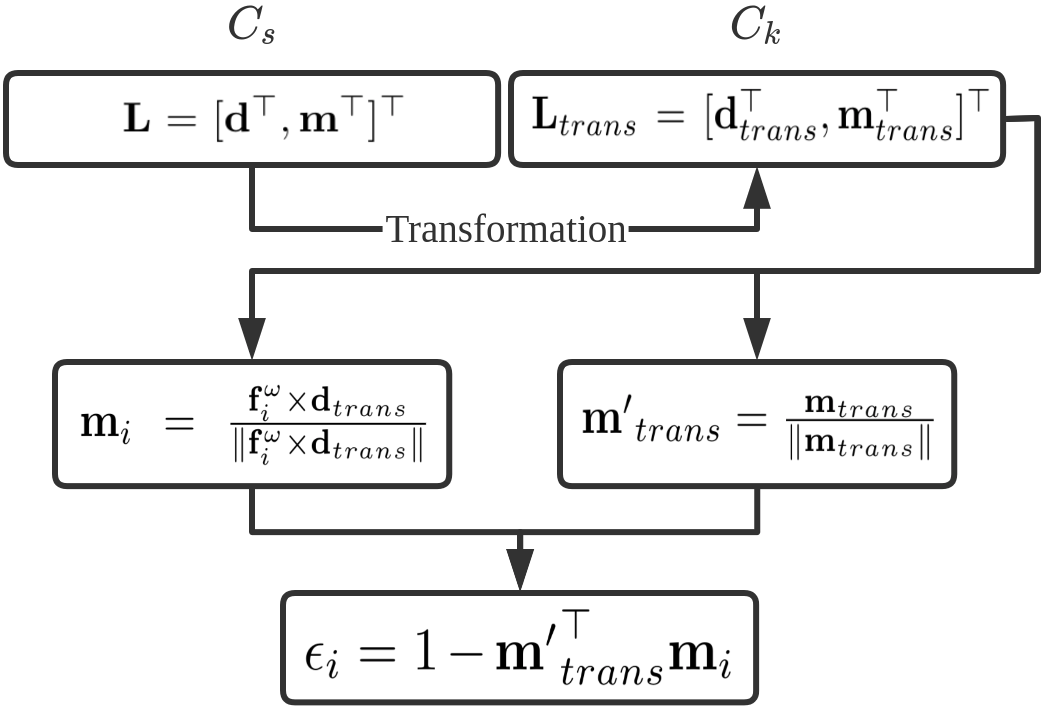}
	\end{center}
	\vspace{-0.1in}
	\caption{Flowchart of the proposed inlier metric.}
	\label{fig:inlierMetric}
\end{figure}

In order to obtain precise lines from RANSAC, the inlier metric is of paramount importance.  We utilize an angular error metric. Without providing exhaustive details, $\mathbf{L} = [\mathbf{d}^\top, \mathbf{m}^\top]^\top$ in $c_s$ can be transformed to $c_i$ using the hypothesized motion parameters, and we denote the transformed line by $\mathbf{L}_{trans} = [\mathbf{d}_{trans}^\top, \mathbf{m}_{trans}^\top]^\top$. $\mathbf{m}_i = \frac{\mathbf{f}_i^{\omega} \times \mathbf{d}_{trans}}{\| \mathbf{f}_i^{\omega} \times \mathbf{d}_{trans} \|}$ should have a small angle with $\mathbf{m'}_{trans} = \frac{\mathbf{m}_{trans}}{\|\mathbf{m}_{trans}\|}$, and we use the error $\epsilon_i = 1 - \mathbf{m'}_{trans}^\top \mathbf{m}_i$. To facilitate understanding, we have drawn a flowchart of this process in Fig.~\ref{fig:inlierMetric}. Note that we also use this metric in order to decide the sense of the velocity direction, which is only determined up-to-scale. If the direction needs to be reversed, it is indicated by line triangulations behind the camera, which in turn can be recognized by wrongly directed moment vectors $\mathbf{m}_{trans}$.

\subsubsection{Efficiency Improvement}
\label{sec:efficiency}
Given there are many events, we improve efficiency by randomly sub-sampling a maximum number of events from each cluster for line regression. The effect of sub-sampling is analyzed in Sec. \ref{sec:synthetic}.

\subsection{Convergence}
\label{sec:convergence}
For each event $e_{ij}$, we reproject each 3D line back to a virtual frame at time $t_{ij}$ and evaluate the orthogonal event-to-line error. Given an inlier threshold, we may thus obtain an inlier ratio $p_{j}$ for each line or cluster. If the estimated $\mathbf{v}$ is accurate, the inlier percentages $p_{j}$ should all be simultaneously at a high level. Therefore, the metric we use to form a termination criterion in the outer layer RANSAC loop is given by

\begin{equation}
   p_{mean} = \frac{\sum_{j=1}^{M} p_j}{M},
\label{equ:p_mean}
\end{equation}
and we terminate the algorithm as soon as this value exceeds a certain threshold $p_{thres}$ or we reach the maximum number of iterations.

\section{Back-end}
\label{sec:back_end_opt}

\begin{figure}[b]
    \centering
    \includegraphics[width=0.9\linewidth]{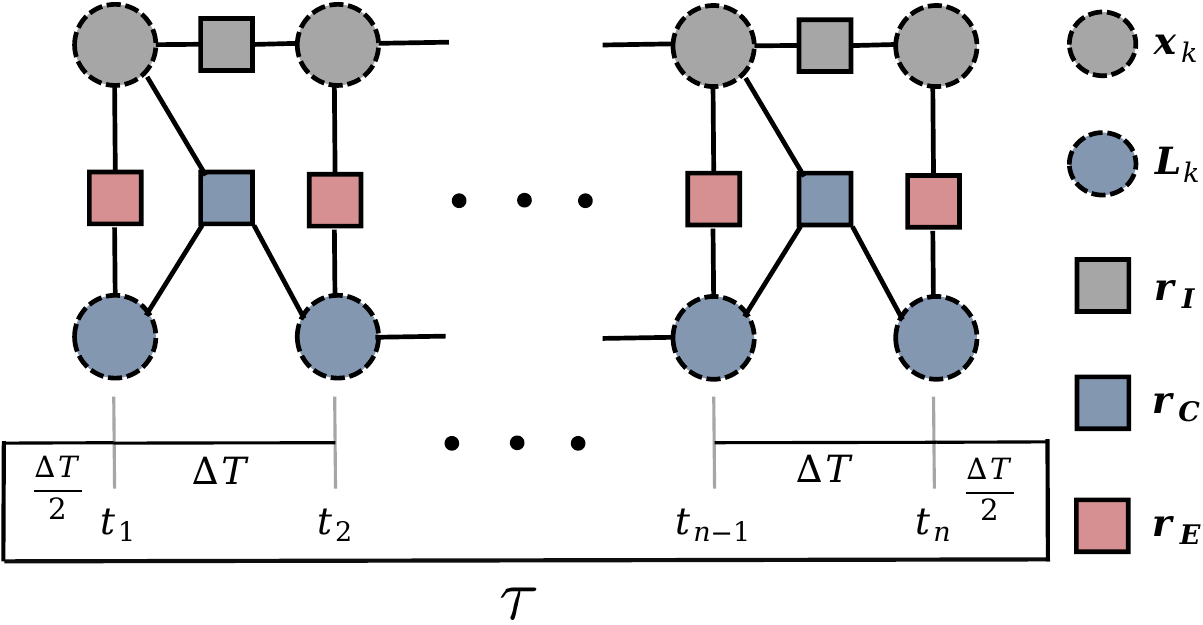}
    \caption{Graphical model and the defined temporal slices of the state variables and measurements involved in the optimization back-end.}
    \label{fig:graph}
\end{figure}

We now proceed to the non-linear optimization back-end, a  sliding-window tightly-coupled monocular visual-inertial velocity estimator. Fig.~\ref{fig:graph} illustrates the factor graph of our sliding window optimizer, where measurements are displayed as square boxes, and estimated variables as circular. The optimizer minimizes the continuous event-line errors over a larger time interval by parameterizing the velocity at multiple points in this interval and regularizing the estimated velocities via IMU preintegration terms.

\subsection{Formulation}
\label{sec:backendformulation}

We denote $\tau$ the length of the time interval over which the integral sliding window stretches. The latter is divided into $n$ contiguous temporal slices of length $\Delta T = \frac{\tau}{n}$. The center time of each temporal slice is denoted by $t_1, \ldots, t_n$, and the IMU frame at $t_k$ is denoted by $\mathcal{F}_k$. The full state vector in the sliding window is defined as

\begin{equation}
\begin{split}
    \mathcal{X} 
    & = [\mathbf{x}_1, \mathbf{x}_2, ... ,\mathbf{x}_n, \mathbf{L}_1, \mathbf{L}_2, ... , \mathbf{L}_m],\\
    \mathbf{x}_k 
    & = [\mathbf{v}_{k}^w, \mathbf{q}_{k}^w, \mathbf{b}_{ak}, \mathbf{b}_{gk}], k = 1, ... , n,
\end{split}
\end{equation}
where $\mathbf{x}_k$ is the dynamic camera state at $t_k$. $\mathbf{x}_k$ contains the velocity $\mathbf{v}_k^w$ in the world frame, the accelerometer and gyroscope biases $\mathbf{b}_{ak}$ and  $\mathbf{b}_{gk}$ expressed in the body frame, and the exterior orientation of the body $\mathbf{q}_{k}^w$ with respect to a gravity-aligned world coordinate system expressed as a quaternion vector. The latter must be estimated in order to properly account for the gravity measurement of the accelerometer readings. $m$ represents the total number of locally estimated line features. Each line $\mathbf{L}_k$ is defined locally within the reference frame of one temporal slice. We represent the lines using Pl\"ucker line coordinates which simplifies geometric line transformations, and we interchangingly employ the orthonormal representation for nonlinear optimization. Note that line features are represented locally for each temporal slice, no global representations are used. As the same line may be observed in multiple subsequent temporal slices, it means that the same line may be represented multiple times across several contiguous temporal slices. We add further regularization terms to avoid strong deviations between such duplicate representations.

Formally, our objective consists of finding the maximum likelihood estimate by minimizing the cost function

\begin{equation}
    \mathcal{X}^* = \underset{\mathcal{X}}{\mathrm{argmin}}~F(\mathcal{X}),
\end{equation}
where
\begin{equation}
    F(\mathcal{X}) = 
    \sum_{\alpha_i \in \mathcal{A}} \rho(\|\mathbf{r}_E^{\alpha_i}\|^2_{\Sigma_{\mathbf{r}_E^{\alpha_i}}}) +
    \sum_{\beta_j \in \mathcal{B}} \|\mathbf{r}_C^{\beta_j}\|^2_{\Sigma_{\mathbf{r}_C^{\beta_j}}} + \sum_{k = 1}^{n - 1} \|\mathbf{r}_I^k\|^2_{\Sigma_{\mathbf{r}_I^k}}.
    \label{eq:main}
\end{equation}
$\rho(s)$ is the Huber loss~\cite{huber1992robust}, $\mathbf{r}_E$ and $\mathbf{r}_I$ are the residuals for the event camera and IMU measurements, respectively, and $\mathbf{r}_C$ is the above-mentioned consistency term for corresponding 3D line representations in different reference frames. $\mathcal{A}$ is the set of all possible associations $\alpha_i=\{ a_i,l_i \}$ between the $a_i$-th event and the $l_i$-th line in the current sliding window. By definition, if the timestamp of the $a_i$-th event $t_{a_i}$ lies within the temporal interval of the $k$-th frame (i.e. $t_{a_i} \in [t_k-\frac{\Delta T}{2};t_k+\frac{\Delta T}{2}]$), the $l_i$-th line $\mathbf{L}_{l_i}$ must be defined in the $k$-th reference frame. The association set $\mathcal{A}$ is simply defined by the cluster slices, all events from one cluster slice will be associated to the same local line representation. $\mathcal{B}$ is the set of all possible line-line tuples $\beta_j=\{l_j,l'_j\}$ such that $\mathbf{L}_{l_j}$ and $\mathbf{L}_{l'_j}$ are corresponding lines defined in two adjacent temporal slices. Again, the set $\mathcal{B}$ is defined by sliced event clusters, as a correspondence $\beta_j$ is defined for each adjacent pair of cluster slices. $\Sigma_{\bullet}$ denotes the error space covariance corresponding to the residual $\bullet$. Details on $\mathbf{r}_E$, $\mathbf{r}_I$ and $\mathbf{r}_C$ are provided in the following. Practically, the optimizer is realized using the Ceres Solver~\cite{Agarwal_Ceres_Solver_2022}.

\subsection{Event Measurement Term}
\label{sec:eventsterm}

Given one $\alpha_i = \{a_i, l_i\}$, the formation of the corresponding reprojection error first requires the 3D line $\mathbf{L}_{l_i}$ to be transformed from its local reference frame to the camera coordinate system at the timestamp of the event $e_{a_i}$. Again, let the local reference frame be the $k$-th coordinate frame within the window, i.e. $t_{a_i} \in [t_k-\frac{\Delta T}{2};t_k+\frac{\Delta T}{2}]$. Denote the instantaneous angular velocity and the unknown instantaneous linear velocity at $t_k$ in $\mathcal{F}_{k}$ as $\boldsymbol{\omega}_k$ and $\mathbf{v}_k$, respectively, where $\mathbf{v}_k = \mathbf{R}^{k}_w(\mathbf{q}_k^w)\mathbf{v}_k^w$. Based on the locally constant velocity assumption, we can easily derive the relative rotation $\mathbf{R}(t_{a_i},t_k)$ and translation $\mathbf{t}(t_{a_i},t_k)$ between the above mentioned two frames. Using (\ref{equ:linetrans}), we obtain the transformed 3D line in the instantaneous body frame at the time of the event $t_{a_i}$. We simply denote this line  $\mathbf{L}$ for the remainder of this section. Since only a single event and a single line are addressed, we furthermore simply use $e$ to refer to $e_{a_i}$, and its pixel coordinates are given by $e_x$, $e_y$.

The reprojection error of the event $e$ is defined as the distance between the event and the projected line, expressed in the normalized image plane of the current camera coordinate system. Event $e$ can be projected onto the normalized image plane by 

\begin{equation}
    \mathbf{e} = 
    \mathbf{K}^{-1}
    \begin{bmatrix}
    e_x\\
    e_y\\
    1
    \end{bmatrix}.
\end{equation}
It is easy to see that the third element of $\mathbf{e}$ must be 1, thus $\mathbf{e}$ is on the normalized image plane.

Furthermore, by \cite{he2018pl} and \cite{zhang2015building}, the projection of a 3D line $\mathbf{L} = [\mathbf{d}^\top, \mathbf{m}^\top]^\top$ is determined by the $\mathbf{m}$ component only, not its direction vector $\mathbf{d}$. Therefore, $\mathbf{L}$ can be projected to the camera image plane by 

\begin{equation}
    \mathbf{l} = \mathcal{K}\mathbf{m},
\end{equation}
where $\mathcal{K} = f_xf_y\mathbf{K}^{-\top}$ is the projection matrix for a line feature.

In our case, when projecting a line to the normalized image plane, $\mathcal{K}$ is an identity matrix. Then, by the formula of the distance between a point and a line in a plane, the distance $d(\mathbf{e}, \mathbf{l})$ between $\mathbf{e}$ and $\mathbf{l}$ is easily deduced as

\begin{equation}
\mathbf{r}_{E} = 
    d(\mathbf{e}, \mathbf{l}) = \frac{\mathbf{e}^\mathrm{T}\mathbf{l}}{\sqrt{l_1^2 + l_2^2}}.
\end{equation}
$l_1$ and $l_2$ here are the first and second elements of $\mathbf{l}$, respectively. Assuming isotropic Gaussian noise on the event position in the image plane, it is fair to assume isotropic Gaussian noise on the normalized event coordinates. Owing to their linear appearance in the residual expression, their propagation into error space is the identity transformation, and hence the covariance reweighting in Eq.~\ref{eq:main} merely appears as a constant factor. Combined with an overall balancing of the different terms in Eq.~\ref{eq:main}, the covariance reweighting of the event measurement term simply becomes a diagonal constant weighting factor.

\subsection{IMU Measurement Term}

Next we explain the IMU measurement terms, which use pre-integrated IMU signals in order to regularize delta-velocities across the window. Common visual-inertial frameworks such as VINS-Mono~\cite{qin18} and OKVIS~\cite{leutenegger2015keyframe} are position-based, and hence require double integration of the IMU signals. As this may lead to fast error accumulation, we conceive it an advantage in our framework that only single signal integrations are required.

The IMU measurements are affected by acceleration bias $\mathbf{b}_a$, gyroscope bias $\mathbf{b}_w$, and additive noise. The raw gyroscope and accelerometer measurements, $\hat{\boldsymbol{\omega}}$ and $\hat{\mathbf{a}}$, are given by
\begin{flalign}
    &\hat{\mathbf{a}}_t = \mathbf{a}_t + \mathbf{b}_{a_t} + \mathbf{R}^t_w\mathbf{g}^w + \mathbf{n}_a, \nonumber \\
    &\hat{\boldsymbol{\omega}}_t = \mathbf{\boldsymbol{\omega}}_t + \mathbf{b}_{\omega_t} + \mathbf{n}_\omega,
\end{flalign}
where $\mathbf{g}^w = [0, 0, g]^\top$ is the gravity vector in the world frame. We assume that the additive noise $\mathbf{n}_a$ and $\mathbf{n}_\omega$ are Gaussian white noise. Acceleration bias and gyroscope bias are modeled as random walk.

As indicated in~\cite{qin18}, for two consecutive temporal slices in $t_k$ and $t_{k+1}$, multiple inertial measurements in $[t_k, t_{k+1}]$ can be integrated in $\mathcal{F}_k$ by

\begin{flalign}
    &\beta^k_{k+1} = \int_{t \in [t_k, t_{k+1}]} \mathbf{R}_t^k(\hat{\mathbf{a}}_t - \mathbf{b}_{a_t})~\mathrm{dt}, \nonumber \\
    &\gamma^k_{k+1} = \int_{t \in [t_k, t_{k+1}]} \frac{1}{2}\Omega(\hat{\boldsymbol{\omega}}_t - \mathbf{b}_{w_t})\gamma^k_{t}~\mathrm{dt},
\end{flalign}
where, 

\begin{flalign}
    &\Omega(\boldsymbol{\omega}) = 
    \left[ \begin{matrix}
    -\lfloor \boldsymbol{\omega} \rfloor_\times & \boldsymbol{\omega}\\
    -\boldsymbol{\omega}^\top & 0
    \end{matrix} 
    \right]
    , 
\end{flalign}
and $\lfloor \boldsymbol{\omega} \rfloor_\times$ is the skew-symmetric matrix of $\boldsymbol{\omega}$. IMU measurements are discrete, and we use the midpoint integration here.

The final IMU pre-integration based residuals become

\scriptsize
\begin{eqnarray}
 \mathbf{r}_{I} & = & 
 \left[ \begin{matrix}
    \delta \boldsymbol{\beta}^{k}_{k+1}\\
    \delta \boldsymbol{\theta}^{k}_{k+1}\\
    \delta \mathbf{b}_a\\
    \delta \mathbf{b}_g
    \end{matrix} 
    \right]\\ & = & 
    \left[ \begin{matrix}

    \mathbf{R}^{k}_w(\mathbf{q}_k)(\mathbf{v}_{k+1}^w + \mathbf{g}^w \Delta T - \mathbf{v}_{k}^w) - \hat{\boldsymbol{\beta}}^{k}_{k+1}\\
    2[{\mathbf{q}_{k}^w}^{-1}\otimes\mathbf{q}_{k+1}^w\otimes(\hat{\gamma}_{k+1}^{k})^{-1}]_{xyz}\\
    \mathbf{b}_{a,k+1} - \mathbf{b}_{a,k}\\
    \mathbf{b}_{w,k+1} - \mathbf{b}_{w,k}
    \end{matrix} 
    \right]\nonumber,
\end{eqnarray}
\normalsize
where $\delta \boldsymbol{\beta}^{k}_{k+1}$, $\delta \boldsymbol{\theta}^{k}_{k+1}$, $\delta \mathbf{b}_a$ and $\delta \mathbf{b}_g$ are IMU measurement residuals for velocity, quaternion orientation, acceleration bias and gyroscope bias, respectively. $[\cdot]_{xyz}$ denotes the extraction of the vector part of the quaternion $\mathbf{q}$, which is used for the representation of the error-state. $\hat{\boldsymbol{\beta}}^{k}_{k+1}$ and $\hat{\gamma}_{k+1}^{k}$ are the preintegrated IMU measurement terms between $t_k$ and $t_{k+1}$. Note that for the propagated residual space uncertainties, the reader is kindly referred to~\cite{qin18}.

\subsection{Consistency Term}

Our final residual is the consistency constraint between corresponding local 3D line representations in adjacent temporal slices. For a $\beta^j = \{l_j,l'_j\}$, let $\mathbf{L}_{l_j}=[\mathbf{d}_{l_j}^\top,\mathbf{m}_{l_j}^\top]^\top$ denote the line in the earlier temporal slice, and $\mathbf{L}_{l'_j}=[\mathbf{d}_{l'_j}^\top,\mathbf{m}_{l'_j}^\top]^\top$ denote the line from the later temporal slice. Let $\mathbf{L}_{l_j}$ furthermore be a line defined in the $k-1$-th local reference frame, and $\mathbf{L}_{l'_j}$ be a line defined in the $k$-th local reference frame. The consistency term is easily formulated by transforming $\mathbf{L}_{l'_j}$ from $\mathcal{F}_{k}$ to $\mathcal{F}_{k-1}$, and evaluating the difference to the line $\mathbf{L}_{l_j}$ originally expressed in $\mathcal{F}_{k-1}$. Similar to Sec.~\ref{sec:eventsterm}, we can obtain the relative rotation $\mathbf{R}_{k-1, k}$ and translation $\mathbf{t}_{k-1, k}$. By Eq.~(\ref{equ:linetrans}), we have

\begin{equation}
    \begin{bmatrix}
    \tilde{\mathbf{m}}_{l_j} \\
    \tilde{\mathbf{d}}_{l_j} 
    \end{bmatrix}
    =
    \begin{bmatrix}
    \mathbf{R}_{k-1, k} & [\mathbf{t}_{k-1, k}]_{\times}\mathbf{R}_{k-1, k} \\
    \mathbf{0} & \mathbf{R}_{k-1, k} 
    \end{bmatrix}
    \begin{bmatrix}
    \mathbf{m}_{l'_j} \\
    \mathbf{d}_{l'_j} 
    \end{bmatrix},
\end{equation}
where $[\tilde{\mathbf{d}}_{l_j}^\top, \tilde{\mathbf{m}}_{l_j}^\top]^\top$ is the predicted Pl\"ucker line at time $t_{k-1}$, and  $[{\mathbf{d}}_{l_j}^\top, {\mathbf{m}}_{l_j}^\top]^\top$ is the optimized Pl\"ucker line at time $t_{k-1}$, The residual expresses the consistency as

\begin{eqnarray}
\begin{scriptsize}
 \mathbf{r}_{C} = 
 \left[ \begin{matrix}
    \delta \mathbf{d}^{k-1}_{j,k}\\
    \delta \mathbf{m}^{k-1}_{j,k}
    \end{matrix} 
    \right] = 
    \left[ \begin{matrix}
    \angle{(\mathbf{d}_{l_j},\tilde{\mathbf{d}}_{l'_j})}\\
    \mathbf{m}_{l_j} - \tilde{\mathbf{m}}_{l'_j}
    \end{matrix} 
    \right],
\end{scriptsize}
\end{eqnarray}
the angle $\angle(\cdot)$ here is measured in radians. Note that for the line consistency term, the covariance reweighting is again formed by employing a constant diagonal matrix of weighting factors. 
However, different weights are used for $\delta \mathbf{d}^{k-1}_{j,k}$ and $\delta \mathbf{m}^{k-1}_{j,k}$ to balance penalties imposed on direction and moment errors, respectively.

\subsection{Further Details}

The bootstrapping algorithm proposed in Sec.~\ref{sec:initialization} is only used at the very beginning of the estimation process. The attentive reader will notice that the scale of the problem is still left uninitialized. While closed-form solutions exist, here we simply propagate a consistent scale factor through-out the window by enforcing line-depth consistencies, and subsequently minimize the IMU measurement term over a single, scalar scale factor. Once initialized, each run of subsequent sliding window optimization is initialized by simply reusing the values of the previous run or---for new data slices---by appending a first-order integration of the inertial measurements to the most recent, already optimized dynamic camera pose. 3D lines are initialized using fast linear line triangulation as explained in the following.

The clustering algorithm of IDOL~\cite{le2020idol} returns the set of events triggered by a moving reprojected line. 
Specifically, the line clustering is continuous and therefore tells us the location of the image plane projection $\mathbf{l}$ of the 3D line $\mathbf{L}$ at any given time. If sufficient displacement has taken place, line triangulation will performed.

As mentioned in Sec. \ref{sec:line_rep}, Pl\"ucker line coordinates are convenient to be used for geometry-related operations. We use them for the initialization of the 3D lines as well. A 3D line can be obtained by the intersection of two planes. Let $\boldsymbol{\pi}_s \in \mathbb{R}^4$ and $\boldsymbol{\pi}_e \in \mathbb{R}^4$ be the observation planes for a 3D line obtained at two different times but already expressed in a common reference frame. The dual Pl\"ucker representation $\mathbf{L}^*$ for a 3D line is formed by the intersecting $\boldsymbol{\pi}_s$ and $\boldsymbol{\pi}_e$~\cite{hartley2003multiple} as in

\begin{equation}
\begin{split}
    \mathbf{L}^*
    & =  \boldsymbol{\pi}_s\boldsymbol{\pi}_e^\top - \boldsymbol{\pi}_e\boldsymbol{\pi}_s^\top \in \mathbb{R}^{4 \times 4}\\
    & = \begin{bmatrix}
    [\mathbf{d}]_{\times} & \mathbf{m} \\
    -\mathbf{m}^\top & 0
    \end{bmatrix}.
\end{split}
\end{equation}

The Pl\"ucker line coordinates $\mathbf{L} = [\mathbf{d}^\top, \mathbf{m}^\top]^\top$ are readily extracted. Note that there are several assumptions made here:
\begin{itemize}
    \item The cluster effectively originates from a moving line in the image.
    \item The planes obtained from lines fitted to events at the beginning and the end of the event cluster are accurate and not too much influenced by noise and outliers.
    \item The line is observed for a sufficiently long time to be able to contribute to the accuracy of the estimation.
\end{itemize}
Those assumptions are generally not satisfied, and we need RANSAC to robustly initialize the lines. However, in practice, our experience has shown that the simple addition of the Huber loss $\rho(s)$ mentioned in Sec. \ref{sec:backendformulation} is sufficient to maintain stable and accurate estimation.

\section{Experiments}
\label{sec:experiment1}

We proceed to the experimental validation of our method. We start with the velocity boot-strapping algorithm, and compare it against the M-estimator method proposed by Peng~et~al.~\cite{peng2021continuous} on both synthetic and real data. Finally, conclude with experimental results over the complete pipeline including sliding-window back-end optimization.

\begin{figure}[b!]
    \centering
    \subfigure[]{
        \includegraphics[width = 0.22\textwidth]{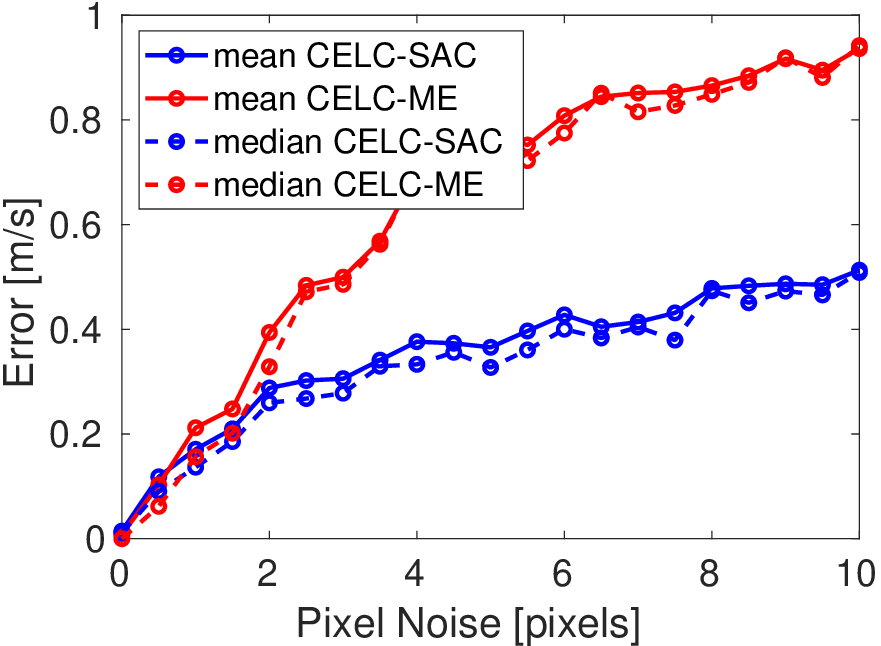}
        \label{fig:eventline_eventNoise}
    }
    \subfigure[]{
    \includegraphics[width = 0.22\textwidth]{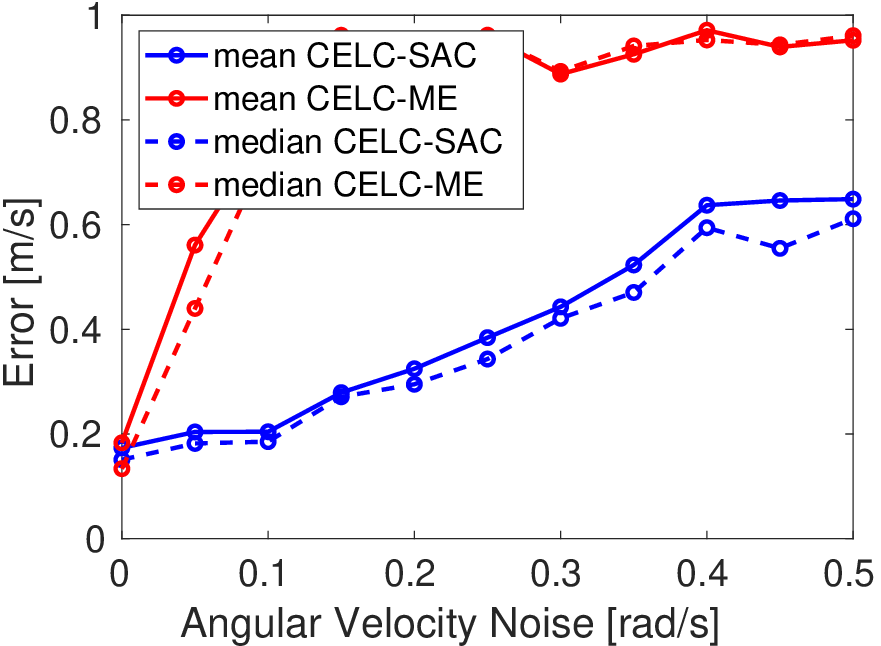}
    \label{fig:eventline_wNoise}
    }
    \caption{Noise analysis. (a) Error as a function of noise on event locations. (b) Errors as a function of different levels of noise added to the angular velocity input. Errors generally increase with noise. Note that CELC-ME represents our previous work~\cite{peng2021continuous}, while CELC-SAC means method presented in this paper.}
    \label{fig:eventline_noise}
    \vspace{-0.05in}
\end{figure}

\begin{figure}[t]
    \centering
    \subfigure[]{
    \includegraphics[width = 0.45\textwidth]{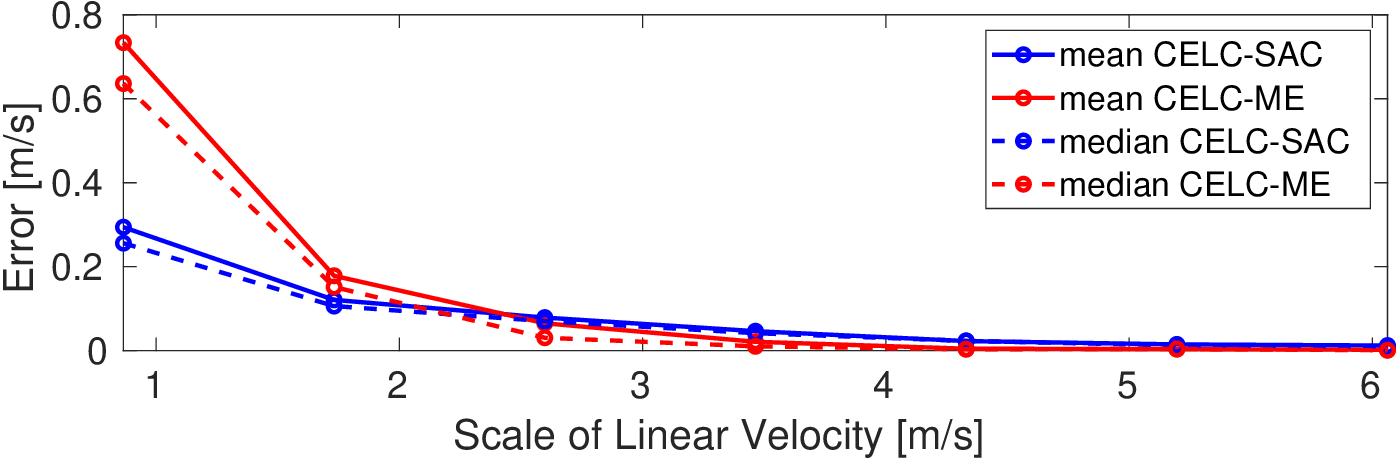}
    \label{fig:eventline_speed}
    }
    \subfigure[]{
    \includegraphics[width = 0.225\textwidth]{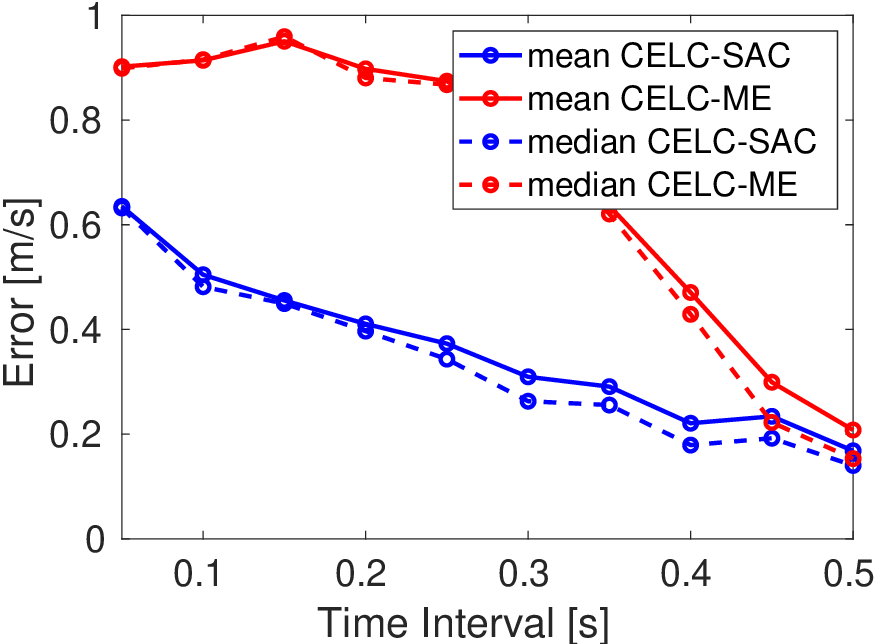}
    \label{fig:eventline_timeInterval}
    }
    \subfigure[]{
    \includegraphics[width = 0.225\textwidth]{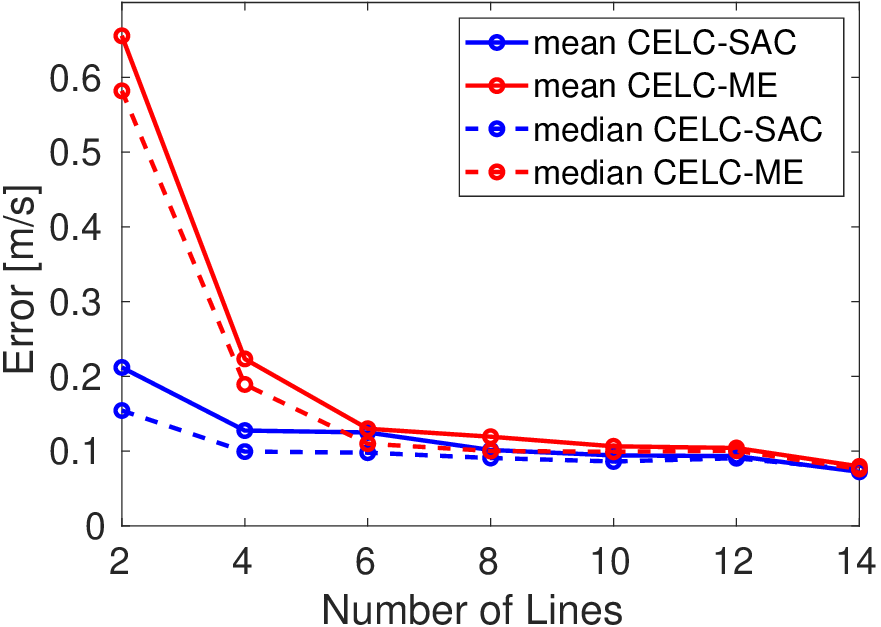}
    \label{fig:eventline_lineNum}
    }
    \subfigure[]{
    \includegraphics[width = 0.225\textwidth]{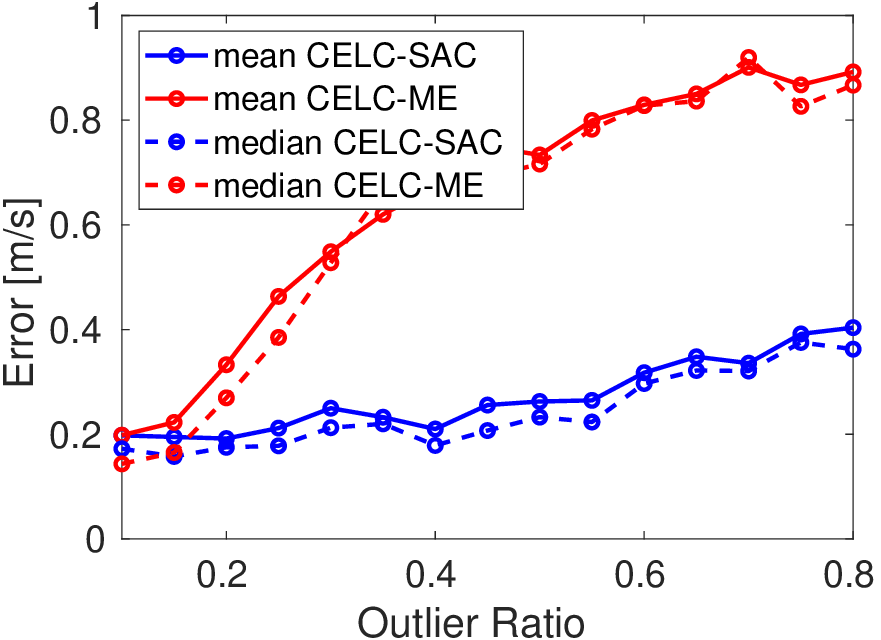}
    \label{fig:eventline_outlierRatio}
    }
    \subfigure[]{
    \includegraphics[width = 0.225\textwidth]{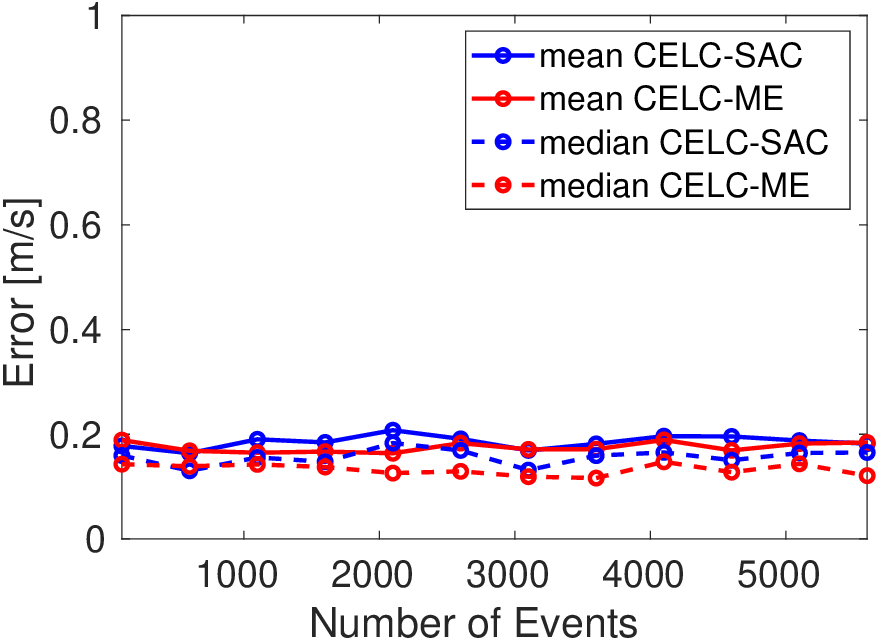}
    \label{fig:eventline_eventsNumRansac}
    }
    \caption{Accuracy for other motion or solver parameters. (a) Errors over an increasing scale of the velocity. (b) Error as a function of the time interval length. (c) Error for different numbers of observed lines. (d) Errors for different outlier ratios. (e) Errors for different numbers of events used within RANSAC.} 
    \label{fig:eventline_effect}
    \vspace{-0.25in}
\end{figure}

\subsection{Synthetic Data Results of the Velocity Initialization}
\label{sec:synthetic}

Here we use our own simulator which generates events based on geometric models (such simulators are common in simulation experiments for traditional geometric vision papers, e.g.~\cite{kneip2011novel, kukelova2008polynomial, yang2015go}). The synthetic data is generated by randomly sampling 5 3D line segments in a cube $([-2,2] \times [-2,2] \times [3,6])$ m$^3$ defined in the camera frame. Next, we randomly generate angular and linear velocities by sampling random vectors $\boldsymbol{\omega}\in[0,1]\times[0,1]\times[0,1]$ rad/s, and random vectors $\mathbf{v}\in[1,1.5]\times[1,1.5]\times[1,1.5]$ m/s. Each event is generated randomly by choosing a 3D point on one of the lines and projecting it into the image plane with the camera pose sampled by a random timestamp within the interval $[0,0.5]$ s. Note that the intrinsic matrix used here is taken from the real data we used in the paper. The pixel location of each generated event is finally disturbed by zero-mean Gaussian noise with a standard deviation of 1 pixel. Besides, we add 10\% outliers to the data. We also add an appropriate amount of Gaussian noise to the ground-truth endpoints of each starting and ending line pair $l_{1j}$ and $l_{3j}$ used in the reference by Peng et al. \cite{peng2021continuous}. This reflects event location uncertainties, event distributions, and the dynamic nature of the line during the time interval $\delta t$. Note that this model does not necessarily adhere to a realistic event generation model but provides basic geometric evaluation cases in which we simply have events generated randomly along the reprojected line.

The experiments analyze the proposed algorithm's performance for variations of both controlled and uncontrolled parameters: i) Evaluation of the solver's robustness against noise, varying speeds, varying outlier ratios, noise on angular velocity, and different numbers of lines; ii) Investigation of design parameter influences such as the interval size and the considered number of events. We evaluate the accuracy by the Euclidean distance $\epsilon$ between the estimated and the ground truth results, which is given by
\begin{equation}
	\epsilon = \|\mathbf{v}_{\text{gt}} - \mathbf{v}_{\text{est}} \|_2, 
\end{equation}
where $\mathbf{v}_{\text{gt}}$ and $\mathbf{v}_{\text{est}}$ are the ground truth and estimated linear velocities, respectively. Note that, as the magnitude of the velocity cannot be found in the monocular case, it is manually set to ground truth. Both the mean and median of $\epsilon$ are indicated for both the proposed solver (CELC-SAC) and the reference implementation of our previous work~\cite{peng2021continuous} (CELC-ME). Results are shown in Fig.~\ref{fig:eventline_noise} and Fig.~\ref{fig:eventline_effect}. 

    \textbf{Robustness against event location noise:} The disturbance of each event is varied with a standard deviation reaching from 0 to 10 pixels.
    As shown in Fig.~\ref{fig:eventline_eventNoise}, our new approach is more robust than our previous work. 

    \textbf{Robustness against noise in the angular velocity input:} The noise is varied between 0 and 0.5 rad/s.
    Fig.~\ref{fig:eventline_wNoise} indicates the corresponding results. It is easy to see that our previous work is very sensitive to the quality of the IMU readings.
    
    \textbf{Effect of linear velocity:} The coordinates of the linear velocity are randomly sampled from the range 
    $[(0.5 + i \times 0.5), (1 + i \times 0.5)]$ m/s, $i = 0,...,6$. The magnitude of the linear velocity is manually reset to $\sqrt{(0.5 + i \times 0.5)^2 \times 3}$.
    The simulation results are shown in Fig.~\ref{fig:eventline_speed}. As can be observed, errors are decreasing with an increasing norm of the speed. Our solver performs better for low velocities, one of the primary weaknesses of the reference algorithm.
    
    \textbf{Effect of the time interval size:} We vary the time interval from 0.05s to 0.5s.
    Results are indicated in Fig.~\ref{fig:eventline_timeInterval} and show that the errors of all solutions are decreasing as the time interval is increasing. However, our newly proposed solver has higher accuracy for small intervals. Note that the constant velocity assumption only holds for small time intervals, which is why stronger performance in this situation is to be regarded as a substantial advantage over the reference implementation.
    
    \textbf{Effect of the number of lines:} The number of lines is varied from 2 to 14.
    Fig.~\ref{fig:eventline_lineNum} presents the results. The accuracy of the solvers increases along with the number of lines. Moreover, our algorithm is more stable when fewer lines are observed.
    
    \textbf{Effect of the outlier ratio:} 
    We vary the outlier ratio from 0.1 to 0.8.
    As can be observed, the proposed solution shows high robustness against increasing outlier ratios (Fig.~\ref{fig:eventline_outlierRatio}).
    
    \textbf{Effect of the number of events:} As detailed in Sec. \ref{sec:efficiency}, the set of events is subsampled in order to increase computational efficiency. The results for different numbers of events is shown in Fig.~\ref{fig:eventline_eventsNumRansac}. 
    The number of events is varied from 100 to 5600.
    Results show that the number of events hardly affects performance.

Overall, the proposed algorithm is more robust against both increasing event location disturbances and higher noise levels in the inertial readings. As demonstrated by our experiments, our new solution achieves higher accuracy than our previous work~\cite{peng2021continuous} when the camera has smaller displacement, a crucial advantage for a direct small time interval-based solution of first-order kinematics.

To compare run times, we used the same settings as those described before. The time interval has been set to 0.3s. The event location noise is set to 0.5 pixels. Additionally, we've set the outlier ratio to 0.1. After 100 runs, our finding is that our algorithm and the algorithm from our previous work have mean run times of 0.0726s and 0.0095s, and median run times of 0.0746s and 0.0089s, respectively. Our algorithm is thus slower, which we trace back to the existence of the inner RANSAC scheme needed to obtain robust estimations of the 3D lines. However, it is important to note that still we take less time than it takes to capture the data, hence the geometric estimation itself can be considered real-time. For reference, Contrast Maximization~\cite{gallego2018unifying}, takes 0.02-0.03s to process 0.01s of event data, which is below real-time capability. It is furthermore possible to optimize parameters by trading accuracy for efficiency (e.g. by limiting the maximum number of iterations), or speeding up execution by making use of parallel computing hardware.

\begin{figure}[t]
    \centering
    {
    \includegraphics[width = 0.115\textwidth]{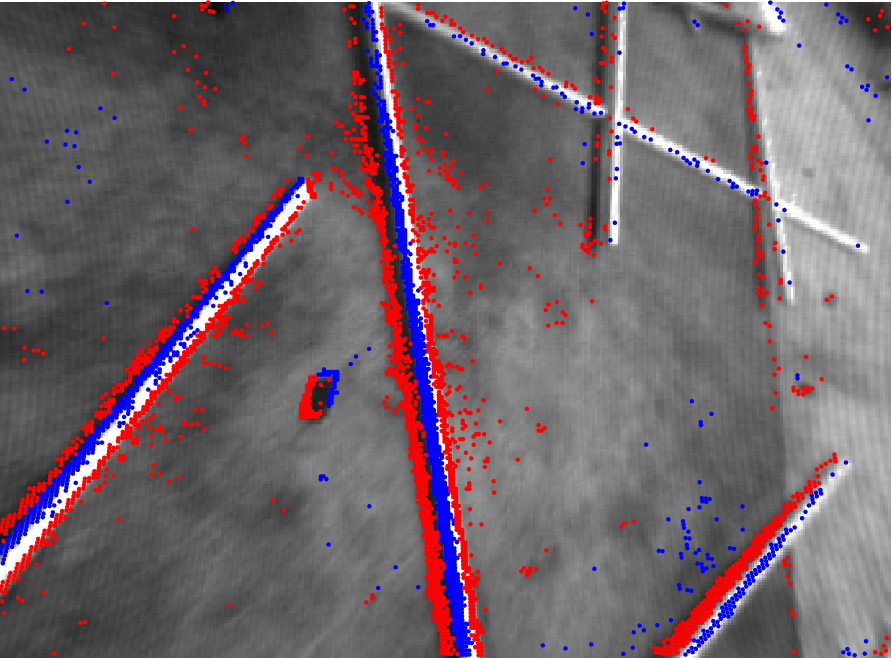}
    \includegraphics[width = 0.115\textwidth]{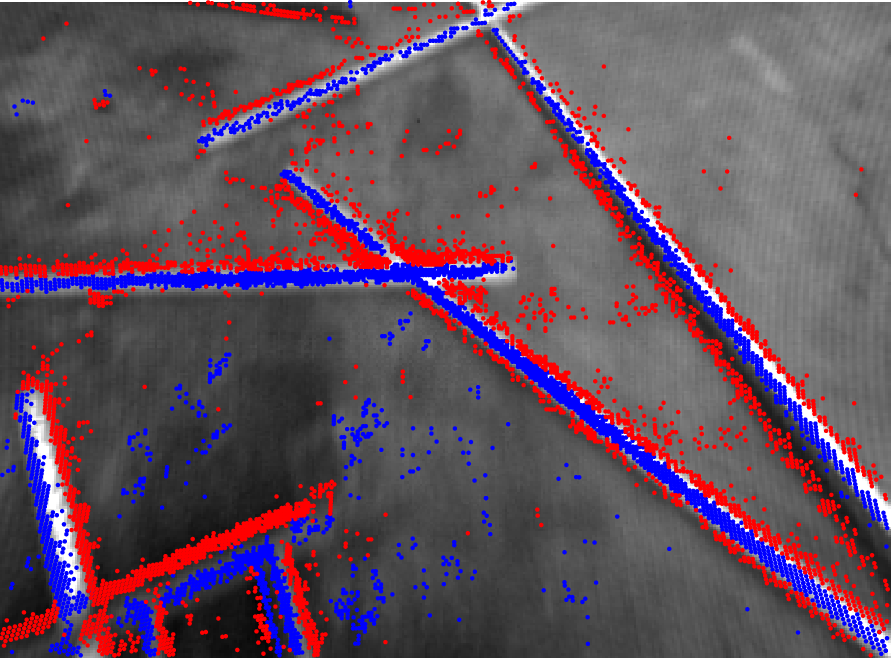}
    \includegraphics[width = 0.115\textwidth]{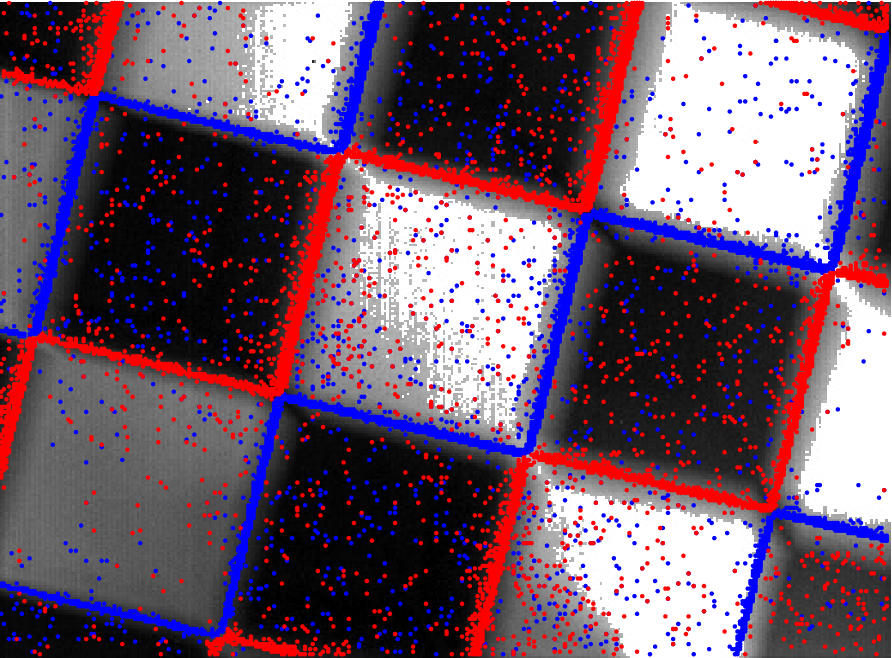}
    \includegraphics[width = 0.115\textwidth]{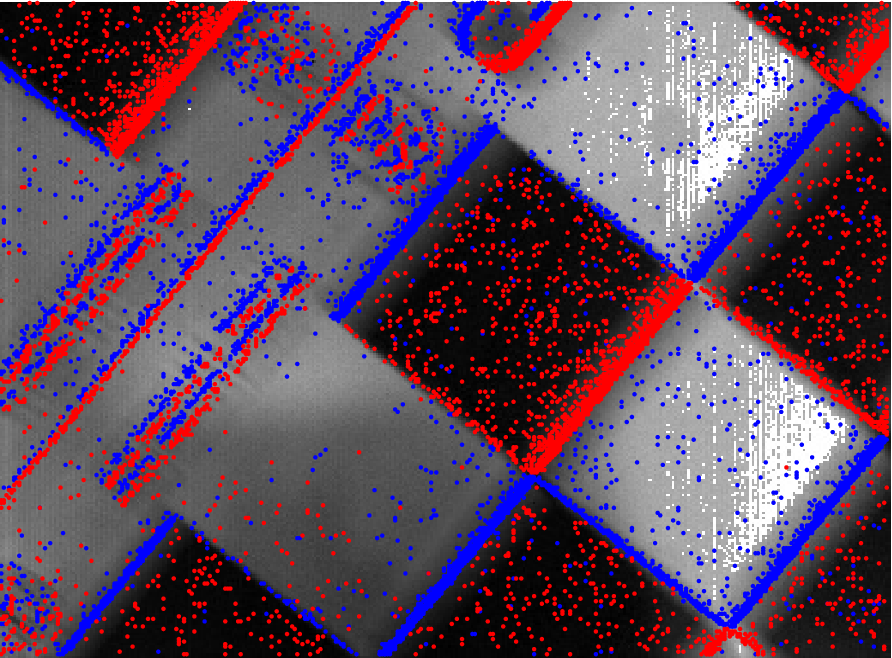}
    }
    \\
    {
    \includegraphics[width = 0.115\textwidth]{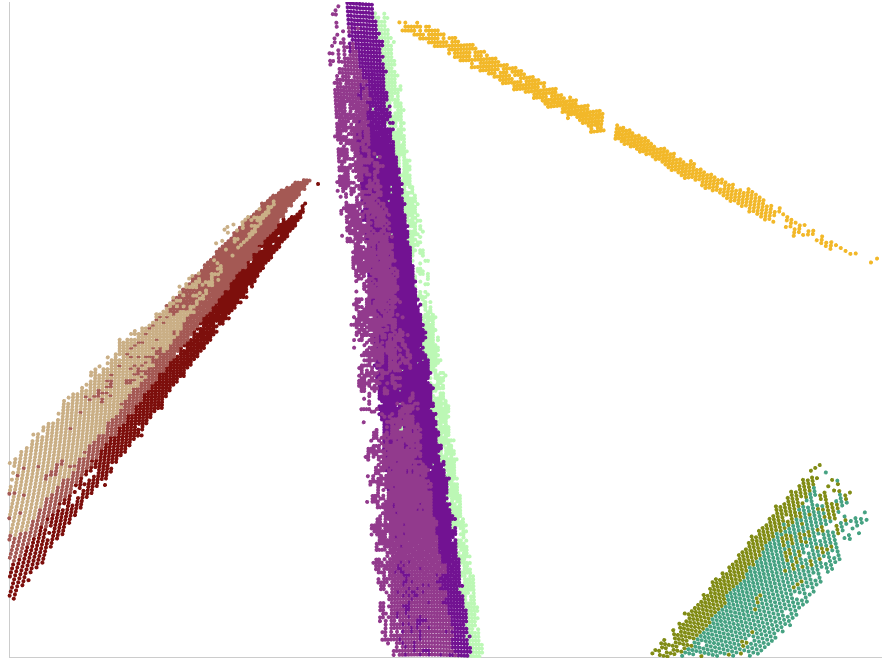}
    \includegraphics[width = 0.115\textwidth]{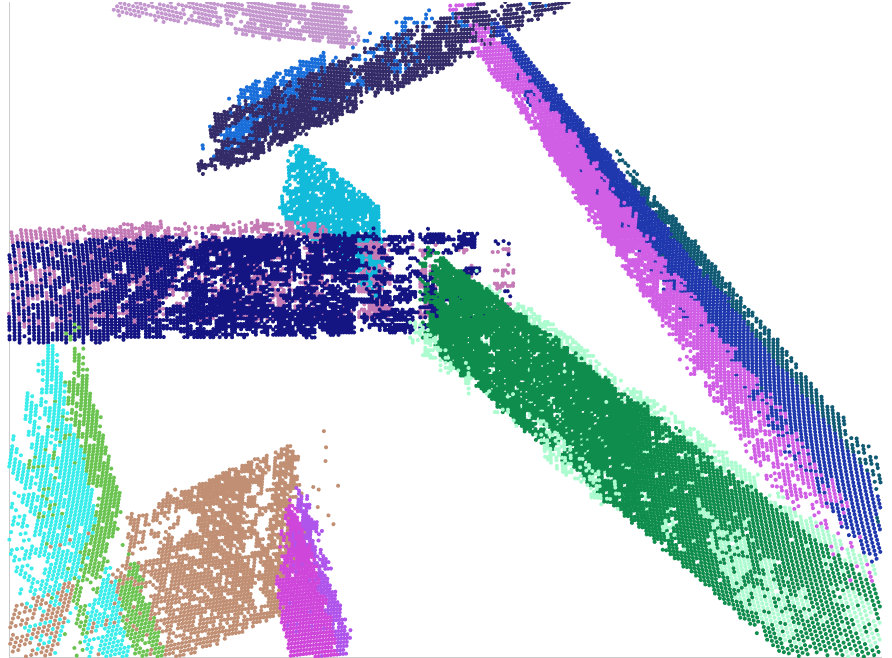}
    \includegraphics[width = 0.115\textwidth]{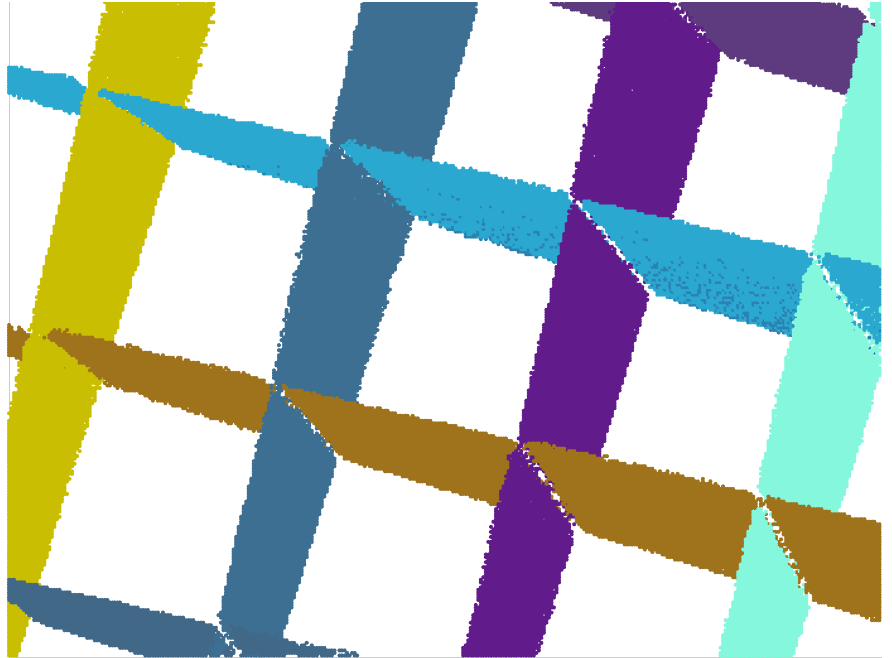}
    \includegraphics[width = 0.115\textwidth]{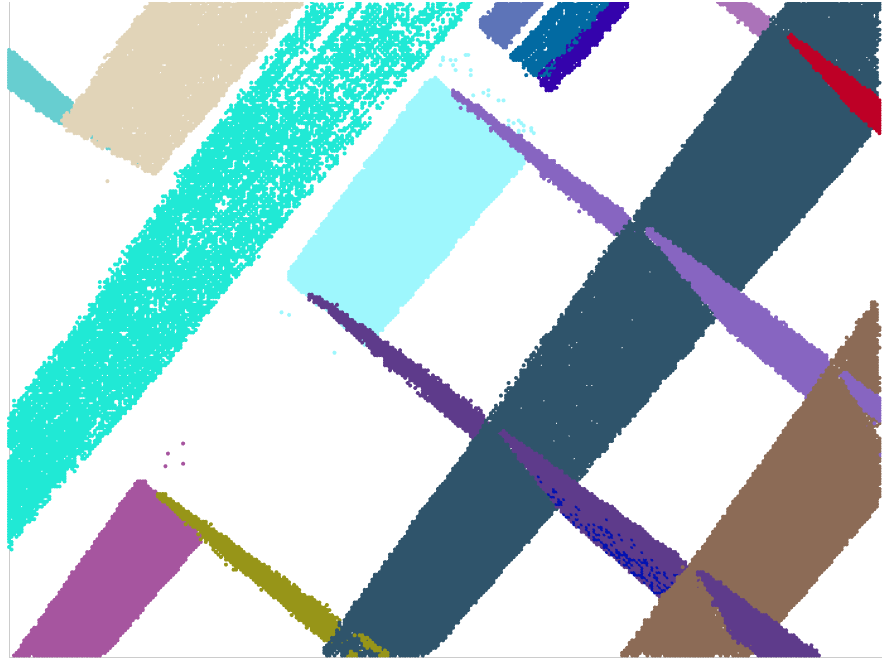}
    }
    \caption{Examples taken from our real data experiments. The first row shows an example grayscale image captured during the time interval (unused) and the corresponding events (positive in blue, negative in red). The second row indicates the identified event clusters corresponding to real-world line segments in a Spatio-temporal view.}
    \label{fig:realImgLine}
\end{figure}

\subsection{Real Data Results of the Velocity Initialization}
\label{sec:real_init}

Next, we evaluate our robust velocity initialization module on real data to verify the practicality of the method. Here we directly evaluate the direction error between ground truth and the estimated linear velocity, which is given by

\begin{equation}
	\theta = \arccos( \frac{\mathbf{v}_{\text{gt}}^{\mathsf{T}} \mathbf{v}_{\text{est}}}{\|\mathbf{v}_{\text{gt}}\|\|\mathbf{v}_{\text{est}}\|}).
\end{equation}

The algorithm is evaluated over two datasets, which are collected by a DAVIS346 event camera with a resolution of 346$\times$260 pixels.  Ground truth is provided by an external motion tracking system. The first sequence is taken from \cite{delmerico2019we} (Indoor45 9), which is captured by an unmanned aerial vehicle (UAV)  carrying a DAVIS346 in a $45^{\circ}$ downward-facing arrangement. The second dataset is collected by a small automated ground vehicle (AGV) and uses a downward facing camera. Examples are presented in Fig.~\ref{fig:realImgLine}.

\begin{figure}[t]
    \centering
    \includegraphics[width = 0.225\textwidth]{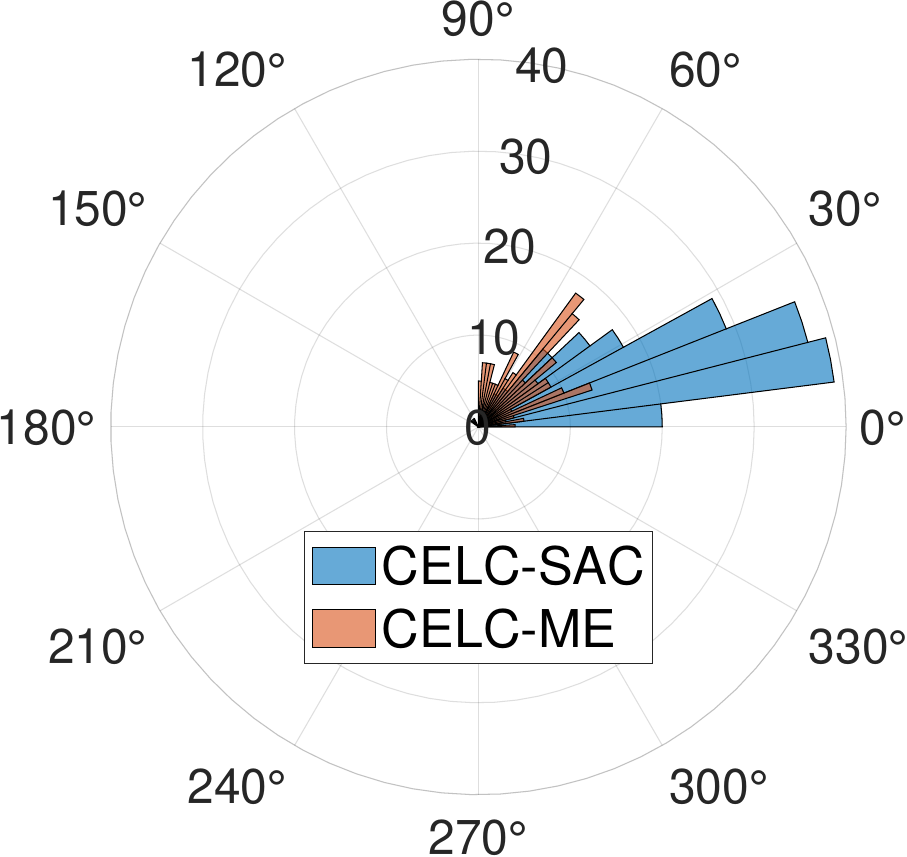}
    \includegraphics[width = 0.225\textwidth]
    {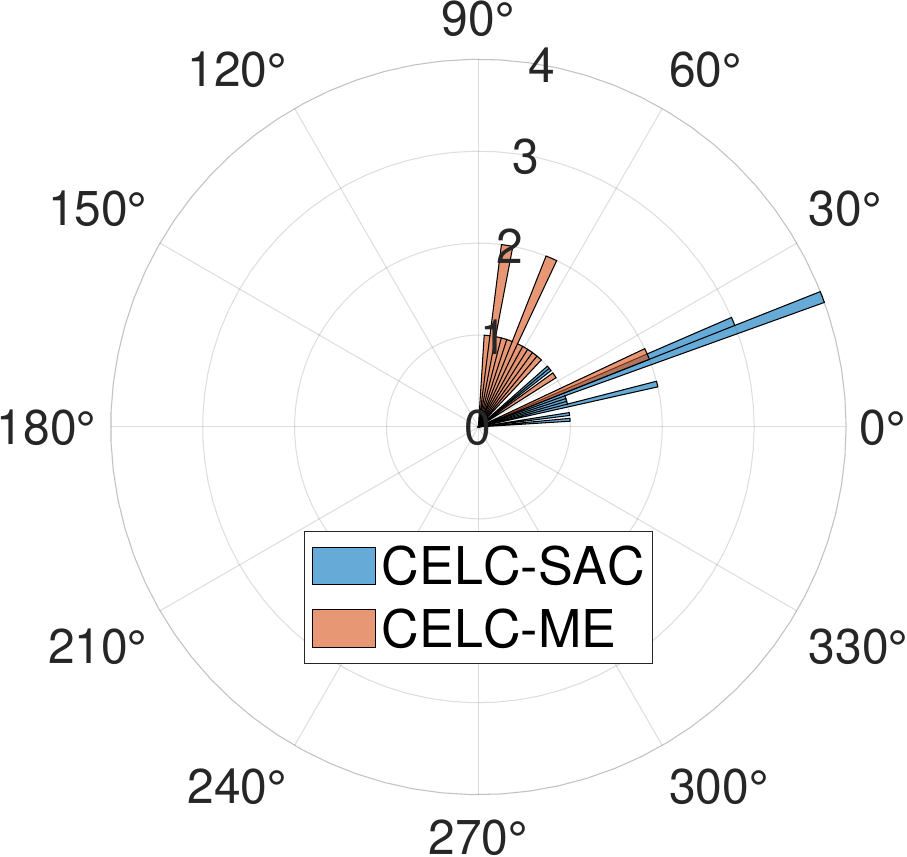}
    \caption{Histogram charts of angular errors in polar coordinates (unit: degrees). Note that CELC-ME represents our previous work~\cite{peng2021continuous}, while CELC-SAC means the method presented in this paper.}
    \label{fig:polarhist}
    \vspace{-0.05in}
\end{figure}

Fig.~\ref{fig:polarhist} shows the histogram charts in polar coordinates of the angle errors $\theta$, and Table \ref{tab:errors} gives the mean and median errors. As can be seen, the distribution of our previous work (CELC-ME) is more scattered, while the distribution of our new approach (CELC-SAC) is more concentrated and has a smaller mean in the distribution region, thus the proposed method obtains more accurate results than our previous work~\cite{peng2021continuous}, which is consistent with the results obtained in simulation. Note that the results of our previous work are worse than the results listed in~\cite{peng2021continuous} given that we use a smaller time interval (about 0.1s vs 0.2s on the UAV dataset, and about 0.4s vs 0.7s on the AGV dataset), and we do not favour our method by choosing smaller time intervals. Rather, smaller time intervals are a requirement because the constant velocity assumption is not valid over longer time intervals. This is why the newly proposed method furthermore succeeds in processing complete sequences rather than only subsets of the data in \cite{peng2021continuous}.

\begin{table}
    \centering
    \caption{Mean ($\alpha$) and Median ($\beta$) angular errors on two datasets from a flying and a ground vehicle platform (unit: rad). }
    \begin{tabular}{c|cccc}
        \toprule
        & \multicolumn{2}{c}{CELC-SAC} & \multicolumn{2}{c}{CELC-ME } \\ 
          & $\alpha(\theta)$    & $\beta(\theta)$  & $\alpha(\theta)$    & $\beta(\theta)$  \\ \midrule
        UAV & 0.4515 & 0.3683 &  0.8214 & 0.8299 \\ 
        AGV  & 0.3517 & 0.3555 & 1.0531 & 1.1141\\
        \bottomrule
    \end{tabular}
    \label{tab:errors}
\end{table}

\label{sec:experiment2}

\subsection{Synthetic Data Results Including Back-end Optimization}

The synthetic data is generated using a modified version of the vio\_data\_simulation library\footnote{\url{https://github.com/HeYijia/vio_data_simulation}}. The original version of the library is used to generate point and line features as measured by traditional cameras. We modified it to generate events triggered by moving line observations.

\begin{figure}[h]
    \centering
    \includegraphics[width=1.0\linewidth]{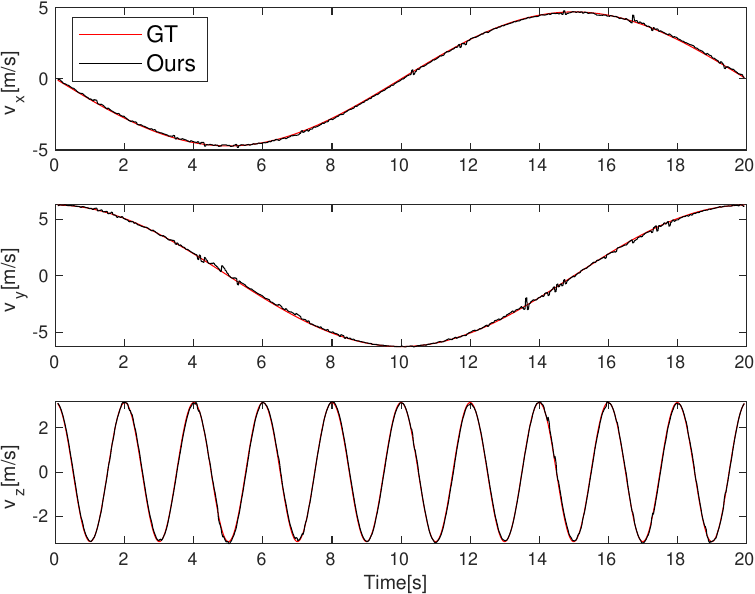}
    \caption{Velocity estimation by the proposed algorithm on synthetic data.}
    \label{fig:vsimu}
\end{figure}

The simulation experiments make use of 10 lines and the size of the sliding window is set to $\tau = 0.1s$. Furthermore, it is assumed that velocities are approximately constant within temporal sub-slices of 0.01s, and hence the overall sliding window is divided into 10 slices. The assumption shows sufficiently validity in practical scenarios, too. The estimated result versus ground truth is shown in Fig.~\ref{fig:vsimu}. Mean and median errors are 0.1365 m/s and 0.1219 m/s, respectively, thus validating the general functionality of the complete framework.

\subsection{Real Data Results Including Back-end Optimization}

We test our algorithm on real sequences of the UZH-FPV Drone Racing dataset~\cite{delmerico2019we}. The dataset is one of the most aggressive visual-inertial odometry dataset to date, and thus represents a significant challenge for state estimation, and a suitable candidate to illustrate the superiority of our algorithm. The UZH-FPV Drone Racing dataset contains many sensors. However, in our experiments, we only use data collected by the event camera (i.e. miniDAVIS346 (mDAVIS)), which provides events, regular frames, and inertial readings. The miniDAVIS346 (mDAVIS) is developed by iniVation\footnote{\url{https://inivation.com/buy/}} and has a spatial resolution of $346 \times 260$ pixels. It generates events with microsecond temporal resolution, and the provided grayscale frames are recorded at 50 Hz. In addition, we used its built-in IMU, which reports inertial readings at 200Hz, and includes data from both a gyroscope and an accelerometer. The dataset contains many sequences. We select three sequences with obvious line features in the scene and with publicly available ground truth, i.e. indoor 45$^{\circ}$ downward facing 2, 4 and 9. Their respective maximum velocities are 6.97 m/s, 6.55 m/s and 11.23 m/s. We compare our work against the open-source frameworks EVO~\cite{rebecq2016evo}, Ultimate SLAM~\cite{vidal2018ultimate}, and VINS-Mono~\cite{qin18}. Given that EVO is a purely vision-based framework not using inertial readings, it requires a good map initialization, it is more suitable for slow scenarios and feature-rich environments. Due to the high dynamics and feature-deprived nature of the FPV sequences (especially in the beginning where there may be short periods with almost no features), EVO will experience tracking loss. Therefore, here we only present the results of Ultimate SLAM and VINS-Mono.

\begin{table*}[]
    \centering
    \caption{Error comparison on different sequences (mean and median error).}
    \begin{tabular}{c|cccc|cccc|cccc}
        \toprule
        \multirow{3}{*}{\bf Sequence} 
        & \multicolumn{4}{c}{Ultimate SLAM (Fr + E + I)} & \multicolumn{4}{c}{VINS} & \multicolumn{4}{c}{Ours} \\ 
        \cline{6-13}
        \cline{2-5}
        & $\mu(\epsilon)$ & $\nu(\epsilon)$ & $\mu(\phi)$ & $\nu(\phi)$ &  $\mu(\epsilon)$ & $\nu(\epsilon)$ & $\mu(\phi)$ & $\nu(\phi)$ & $\mu(\epsilon)$ & $\nu(\epsilon)$ & $\mu(\phi)$ & $\nu(\phi)$ \\ 
        \cline{6-13}
        \cline{2-5}
        &  [m/s] & [m/s] & [1] & [1]  & [m/s] & [m/s] & [1] & [1] & [m/s] & [m/s] & [1] & [1]\\ \midrule
        Seq. 2 & 1.0383 & 1.0436 & 0.2846 & 0.2394 & 0.4683 & 0.4788 & 0.1550 & 0.1162 & {\bf 0.4046} & {\bf 0.3563} & {\bf 0.0927} & {\bf 0.0794} \\ 
        Seq. 4  & 0.8801 & 0.7462 & 0.2637 & 0.2062 & 0.5709 & 0.5771 & 0.2407 & 0.1656 &{\bf 0.3468} & {\bf 0.2985} & {\bf 0.0868} & {\bf 0.0717}\\
        Seq. 9  & 1.9077 & 1.8864 & 0.3097 & 0.3098 & 1.8285 & 1.7943 & 0.3105 & 0.3135 & {\bf 0.6148} & {\bf 0.5327} & {\bf 0.1036} & {\bf 0.0946}\\
        \bottomrule
    \end{tabular}
    \label{tab:errors2}
\end{table*}

\begin{table*}[]
    \centering
    \caption{Error comparison on different sequences (standard deviation and maximum error).}
    \begin{tabular}{c|cccc|cccc|cccc}
        \toprule
        \multirow{3}{*}{\bf Sequence} 
        & \multicolumn{4}{c}{{Ultimate SLAM (Fr + E + I)}} & \multicolumn{4}{c}{VINS} & \multicolumn{4}{c}{Ours} \\ 
        \cline{6-13}
        \cline{2-5}
        & {$\mathrm{std}(\epsilon)$} & {$\mathrm{max}(\epsilon)$} & {$\mathrm{std}(\phi)$} & {$\mathrm{max}(\phi)$} &  $\mathrm{std}(\epsilon)$ & $\mathrm{max}(\epsilon)$ & $\mathrm{std}(\phi)$ & $\mathrm{max}(\phi)$ & $\mathrm{std}(\epsilon)$ & $\mathrm{max}(\epsilon)$ & $\mathrm{std}(\phi)$ & $\mathrm{max}(\phi)$ \\ 
        \cline{6-13}
        \cline{2-5}
        &  {[m/s]} & {[m/s]} & {[1]} & {[1]}  & [m/s] & [m/s] & [1] & [1] & [m/s] & [m/s] & [1] & [1]\\ \midrule
        Seq. 2 & {0.5244} & {2.0186} & {0.1914} & {1.2596} & {\bf 0.2163} & {\bf 1.5358} & {\bf 0.0433} & 1.0894 &  0.2371 &  1.6830 &  0.0564 & {\bf 0.3543} \\ 
        Seq. 4  & {0.4552} & {1.8947} & {0.1156} & {1.0011} & 0.3008 & {\bf 1.1049} & {\bf 0.0573} & 1.0296 &{\bf 0.2308} &  2.1309 &  0.0726 & 0.8426\\
        Seq. 9  & {0.6679} & {3.5409} & {0.0609} & {0.5336} & 0.6678 & 4.7264 & 0.0702 & 0.8853 & {\bf 0.3695} & {\bf 2.4637} & {\bf 0.0526} & {\bf 0.3218}\\
        \bottomrule
    \end{tabular}
    \label{tab:errors3}
\end{table*}

\begin{figure*}[t]
    \centering
    \includegraphics[width = 0.325\textwidth]{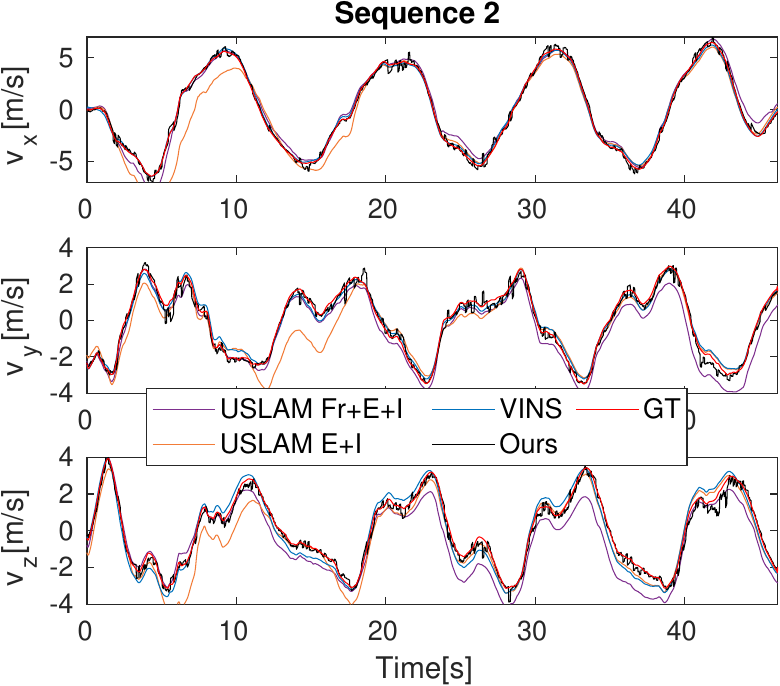}
    \includegraphics[width = 0.325\textwidth]{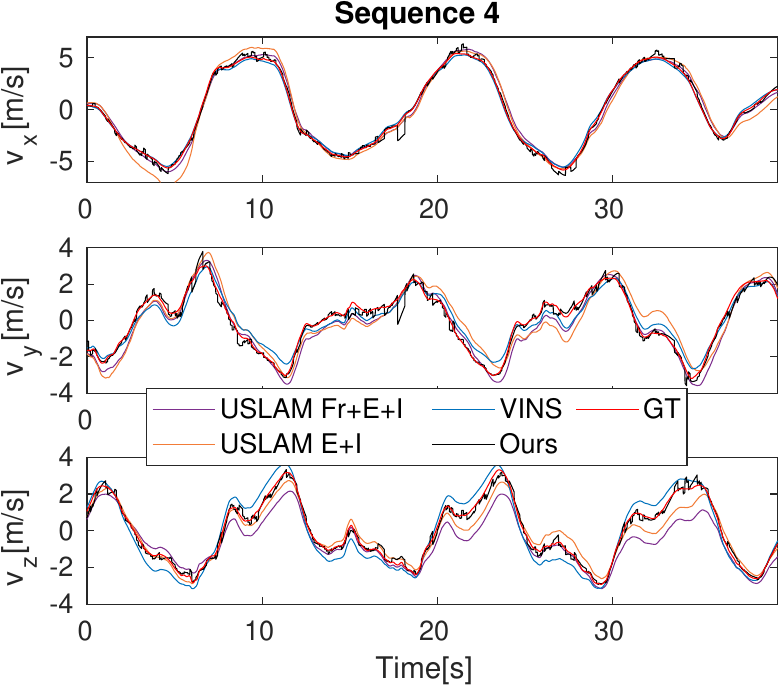}
    \includegraphics[width = 0.325\textwidth]{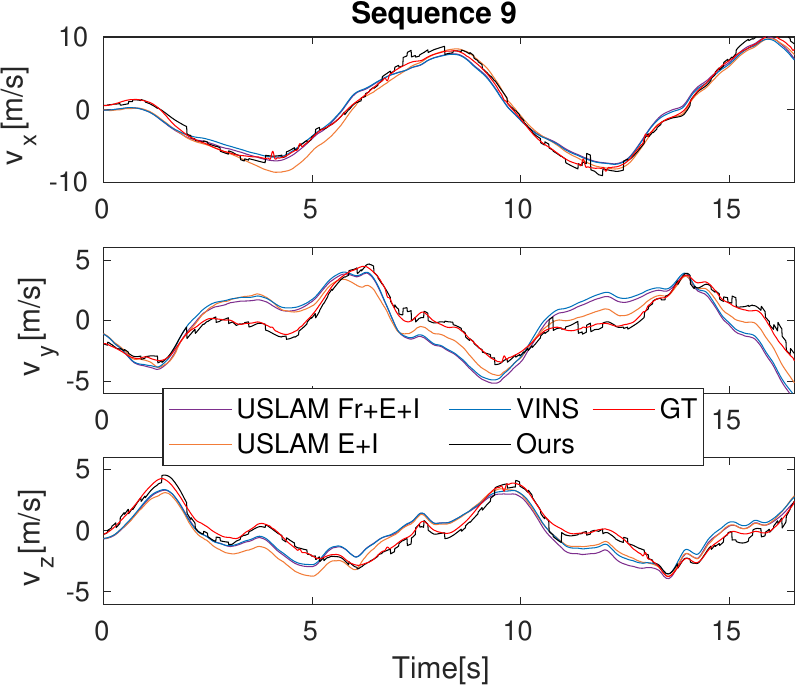}
    \caption{Estimated linear velocities compared against ground truth and VINS results on sequences.}
    \label{fig:fpv}
\end{figure*}

\begin{table}[t]
    \centering
    \caption{Error results of Ultimate SLAM (E + I) on different sequences (mean and median error).}
    \begin{tabular}{c|cccccccc}
        \toprule
        \multirow{3}{*}{\bf Sequence} 
        & \multicolumn{4}{c}{Ultimate SLAM (E + I)} \\ 
        \cline{2-5}
        &  $\mu(\epsilon)$ & $\nu(\epsilon)$ & $\mu(\phi)$ & $\nu(\phi)$  \\ 
        \cline{2-5}
        & [m/s] & [m/s] & [1] & [1]\\ \midrule
        Seq. 2 & 1.2343 & 0.6956 & 0.2848 & 0.1579 \\ 
        Seq. 4  & 1.1104 & 1.1088 & 0.2659 & 0.2617\\
        Seq. 9  & 1.5825 & 1.5014 & 0.2593 & 0.2288\\
        \bottomrule
    \end{tabular}
    \label{tab:err_uslam_e}
\end{table}

\begin{table}[t]
    \centering
    \caption{Error results of Ultimate SLAM (E + I) on different sequences (standard deviation and the maximum error)}
    \begin{tabular}{c|cccccccc}
        \toprule
        \multirow{3}{*}{\bf Sequence} 
        & \multicolumn{4}{c}{Ultimate SLAM (E + I)} \\ 
        \cline{2-5}
        &  $\mathrm{std}(\epsilon)$ & $\mathrm{max}(\epsilon)$ & $\mathrm{std}(\phi)$ & $\mathrm{max}(\phi)$  \\ 
        \cline{2-5}
        & [m/s] & [m/s] & [1] & [1]\\ \midrule
        Seq. 2 & 1.0920 & 5.1026 & 0.2756 & 1.8231 \\ 
        Seq. 4  & 0.3843 & 2.0355 & 0.0619 & {\bf 0.6150}\\
        Seq. 9  & 0.6883 & 4.4225 & 0.1218 & 0.8410\\
        \bottomrule
    \end{tabular}
    \label{tab:err_uslam_e_max}
\end{table}

The sliding window is set to $\tau=0.1s$ and the length of the temporal sub-slices is set to $0.01s$, which is similar to the settings in the synthetic experiment. We analyze our algorithm in one of two ways. The first one is given by error statistics, the second one by a display of the estimated results together with the velocity estimation obtained by VINS and ground truth. Regarding error statistics, we use two different error metrics. The first is the absolute error, which is given by
\begin{equation}
    \epsilon = \|\mathbf{v}_{gt} - \mathbf{v}_{est}\|_2,
\end{equation}
where $\mathbf{v}_{gt}$ and $\mathbf{v}_{est}$ are the ground truth and estimated linear velocities, respectively. The second error metric is the relative error, which is given by

\begin{equation}
    \phi = \frac{\|\mathbf{v}_{gt} - \mathbf{v}_{est}\|_2}{\|\mathbf{v}_{gt}\|_2}.
\end{equation}

\begin{figure}[h]
    \centering
    \includegraphics[width=0.8\linewidth]{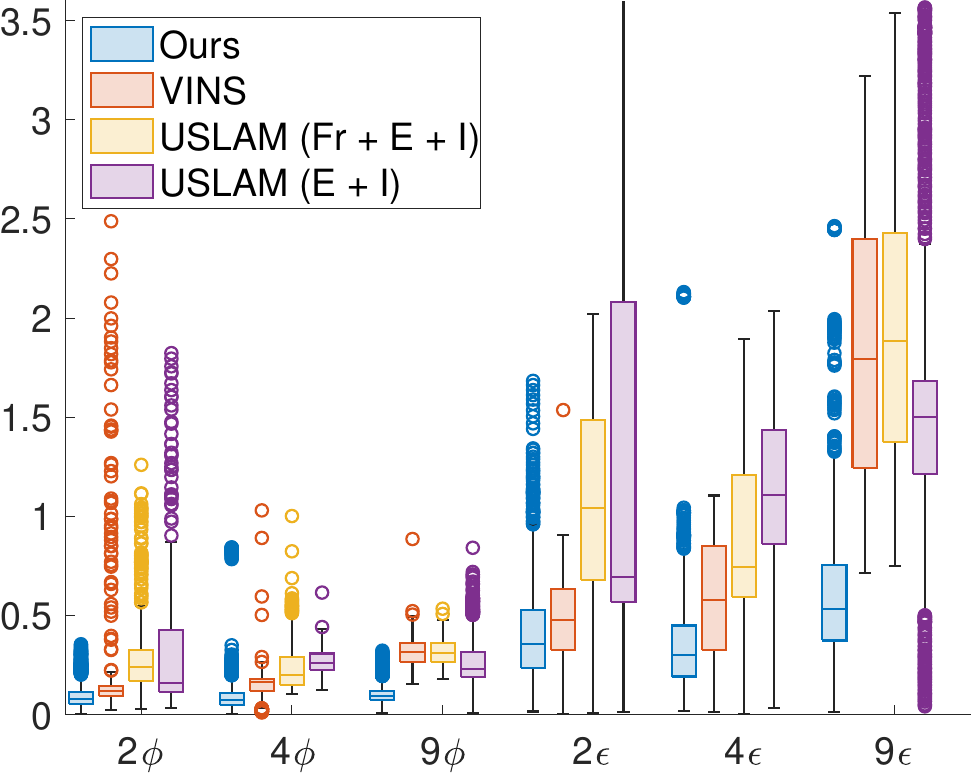}
    \caption{Box plot of error distribution on distinct sequences. The numbers here represent different sequences, while $\phi$ and $\epsilon$ represent different error metrics.}
    \label{fig:distributions}
\end{figure}

Ultimate SLAM has two modes: events-only mode (E + I) and mode including all sensors (Fr + E + I). The comparison results against Ultimate SLAM and VINS on different sequences are presented in Table \ref{tab:errors2} and Table \ref{tab:errors3}, with their box plot error distribution shown in Fig.~\ref{fig:distributions}, ($\mu$ denotes the mean error, $\nu$ the median error, $\mathrm{std}$ the standard deviation, and $\mathrm{max}$ the maximum error). Note that due to space limitations, we have placed the results of Ultimate SLAM (E + I) in Table \ref{tab:err_uslam_e} and Table \ref{tab:err_uslam_e_max}. As can be observed, our estimation gives the best results under all metrics, except for the $\mathrm{max}$ metric of sequence 4, where Ultimate SLAM (E + I) achieved the best results. This is consistent with the results in Fig. \ref{fig:fpv}. As can be observed, the estimated results are all closer to ground truth (GT) than the velocities obtained by Ultimate SLAM and VINS, especially in the last sequence in which the most challenging dynamics occur. The jumps in our results are due to the fact that there are some scenes in the sequences with very few line features. The specific error reduction rate compared against Ultimate SLAM and VINS for all sequences is provided in Table~\ref{tab:rate_uslam}, \ref{tab:rate_uslam_e} and \ref{tab:rate}.

\begin{table}[t]
    \centering
    \caption{Basic properties of the compared sequences.}
    \begin{tabular}{c|cccc}
        \toprule
        \multirow{2}{*}{\bf Sequence} 
        & \multirow{1}{*}{Duration} & \multirow{1}{*}{Length} & \multirow{1}{*}{Max. Speed} & \multirow{1}{*}{Med. Speed} \\ 
        \cline{2-5}
        & [s] & [m] & [m/s] & [m/s] \\ \midrule
        Seq. 2 & 55.77 & 218.90 & 6.97 & 4.45  \\ 
        Seq. 4  & 47.36 & 168.06 & 6.55 & 4.22 \\
        Seq. 9  & 40.00 & 215.58 & 11.23 & 5.66 \\
        \bottomrule
    \end{tabular}
    \label{tab:seqs}
\end{table}

\begin{table}[t]
    \centering
    \caption{Error reduction rate of our method compared to Ultimate SLAM (Fr + E + I).}
    \begin{tabular}{c|cccccccc}
        \toprule
        \multirow{3}{*}{\bf Sequence} 
        & \multicolumn{4}{c}{Reduction Rate} \\ 
        \cline{2-5}
        &  $\mu(\epsilon)$ & $\nu(\epsilon)$ & $\mu(\phi)$ & $\nu(\phi)$  \\ 
        \cline{2-5}
        & [\%] & [\%] & [\%] & [\%]\\ \midrule
        Seq. 2 & 61.04 & 65.86 & 67.45 & 66.85 \\ 
        Seq. 4  & 60.59 & 59.99 & 67.07 & 65.22\\
        Seq. 9  & 67.77 & 71.76 & 66.54 & 69.48\\
        \bottomrule
    \end{tabular}
    \label{tab:rate_uslam}
\end{table}

\begin{table}[t]
    \centering
    \caption{Error reduction rate of our method compared to Ultimate SLAM (E + I).}
    \begin{tabular}{c|cccccccc}
        \toprule
        \multirow{3}{*}{\bf Sequence} 
        & \multicolumn{4}{c}{Reduction Rate} \\ 
        \cline{2-5}
        &  $\mu(\epsilon)$ & $\nu(\epsilon)$ & $\mu(\phi)$ & $\nu(\phi)$  \\ 
        \cline{2-5}
        & [\%] & [\%] & [\%] & [\%]\\ \midrule
        Seq. 2 & {67.22} & {48.77} & {67.47} & {49.75} \\ 
        Seq. 4  & 68.76 & 73.08 & 67.35 & 72.60\\
        Seq. 9  & 61.15 & 64.52 & 60.04 & 58.67\\
        \bottomrule
    \end{tabular}
    \label{tab:rate_uslam_e}
\end{table}

\begin{table}[t!]
    \centering
    \caption{Error reduction rate of our method compared to VINS.}
    \begin{tabular}{c|cccccccc}
        \toprule
        \multirow{3}{*}{\bf Sequence} 
        & \multicolumn{4}{c}{Reduction Rate} \\ 
        \cline{2-5}
        &  $\mu(\epsilon)$ & $\nu(\epsilon)$ & $\mu(\phi)$ & $\nu(\phi)$  \\ 
        \cline{2-5}
        & [\%] & [\%] & [\%] & [\%]\\ \midrule
        Seq. 2 & 13.62 & 25.58 & 40.22 & 31.71 \\ 
        Seq. 4  & 39.24 & 48.28 & 63.93 & 56.71\\
        Seq. 9  & 66.37 & 70.31 & 66.62 & 69.84\\
        \bottomrule
    \end{tabular}
    \label{tab:rate}
\end{table}

\begin{figure}[t]
    \centering
    \subfigure[]
    {
    \includegraphics[width = 0.14\textwidth]{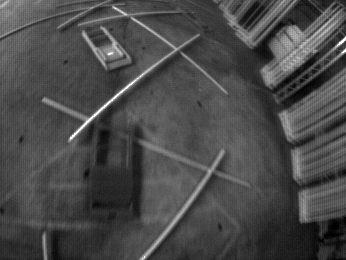}
    \label{fig:fpv2}
    }
    \subfigure[]{
    \includegraphics[width = 0.14\textwidth]{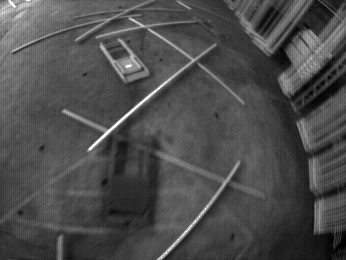}
    \label{fig:fpv4}
    }
    \subfigure[]{
    \includegraphics[width = 0.14\textwidth]{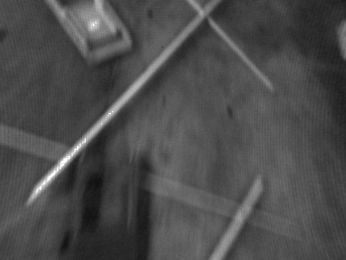}
    \label{fig:fpv9}
    }
    \caption{Comparison of grayscale frames while passing through the same scene in the three sequences. (a), (b) and (c) are from indoor 45$^{\circ}$ downward facing 2, 4 and 9 respectively.}
    \label{fig:imgs}
\end{figure}

The superior performance on the final dataset is grounded on two facts. The first one is the difference between traditional and event cameras. The elevated overall speed in sequence 9 is faster than in sequences 2 and 4 (as illustrated in Table \ref{tab:seqs}), which causes an elevated amount of motion blur in the regular images. For example, Fig.~\ref{fig:imgs} shows a selection of images captured in the same turns in three different sequences. The images taken from sequences 2 and 4 are relatively clear, while the image taken from sequence 9 has more motion blur. These images are indeed representative for the average image quality along the sequences. While motion blur easily affects frame-based algorithms such as VINS, and Ultimate SLAM (Fr + E + I) also uses images. This can also explain why Ultimate SLAM (E + I) achieved the best in the $\mathrm{max}$ metric for sequence 9. Our event-based solution indeed benefits from faster motion as this leads to larger, more easily observable line displacements during small time intervals, and thus better noise cancellation effects. The second argument in support of our newly proposed method is given by differences in the estimated state variables and consistency terms. Our method does not aim at absolute position estimation, hence first-order integrations of inertial signals are sufficient to formulate our optimization problem. This is reflected in our estimation results for sequences 2 and 4, which both reflect higher accuracy for the body-centric, event-inertial approach. 

\subsection{Ablation Study for the Consistency Term}
We also conducted an ablation study on the consistency term, to see if the consistency term can improve the system’s robustness. For this reason, we optimize without consistency term on sequences 2, 4, and 9, and summarize the results in Table~\ref{tab:ablation}. As can be seen, compared against our results from Table~\ref{tab:errors2}, disabling the consistency term simply brakes the framework. We thus verify the effectiveness of the proposed consistency term and the validity of the proposed system design.

\begin{table}[t]
    \centering
    \caption{Results without consistency term}
    \begin{tabular}{c|cccccccc}
        \toprule
        \multirow{3}{*}{\bf Sequence} 
        & \multicolumn{4}{c}{Results Without Consistency Term} \\ 
        \cline{2-5}
        &  $\mu(\epsilon)$ & $\nu(\epsilon)$ & $\mu(\phi)$ & $\nu(\phi)$  \\ 
        \cline{2-5}
        & [m/s] & [m/s] & [1] & [1]\\ \midrule
        Seq. 2 & 26.5778 & 27.1131 & 6.2653 & 5.7262 \\ 
        Seq. 4  & 23.9810 & 22.7299 & 6.1819 & 5.0613\\
        Seq. 9  & 8.1640 & 7.1972 & 1.3389 & 1.1825\\
        \bottomrule
    \end{tabular}
    \label{tab:ablation}
\end{table}

\subsection{Runtime Analysis}

As mentioned above, the two-layer RANSAC algorithm could be considered real-time and further pushed in computational efficiency by optimizing and tuning some of the RANSAC parameters. Furthermore, we believe that it would certainly be possible to make use of parallel computing techniques. Currently, the main obstacle towards real-time processing is given by our front-end line tracking method which is relatively slow, as we utilize a point cloud processing method. It could again be sped up by considering parallelization or using other line tracking methods.

In regards to back-end optimization---and compared against other frameworks such as Ultimate SLAM~\cite{vidal2018ultimate}---it is worth mentioning that one of the main reasons why our algorithm is slower is because we make use of all the events. Ultimate SLAM primarily focuses on creating virtual frames (event frames) from spatiotemporal event windows. It then carries out feature detection and tracking through traditional computer vision techniques, specifically using the FAST corner detector~\cite{rosten2006machine} and the Lucas-Kanade tracker~\cite{lucas1981iterative}. This sparsifies data, and thus leads to much faster execution times. We thus believe that our implementation could also be sped up by considering downsampling techniques.
\vspace{0.1in}
\section{Conclusion}

This paper introduces a novel method for visual-inertial fusion which estimates directly the velocity of the sensor system rather than its global position. This is interesting as it enables fail-safe motion control without depending on the construction or tracking of a globally consistent map. Another natural benefit of fusion at the velocity level is given by the fact that it only requires single integration of inertial readings. Our main novelty consists of the use of an event camera instead of a regular frame-based camera. We demonstrate that this is not only a highly intuitive choice for a dynamic vision sensor, but also beneficial one. As shown, velocity samples can be directly calculated from thin slices of the space-time volume of events. Furthermore, our method indeed keeps performing well in highly dynamic scenarios where regular camera-based approaches fail. Our future work consists of extending the approach from line features to a combination of line and point features. We also consider the addition of a global position and map estimation layer around the presently proposed velocity estimation framework. Our current version is not real-time as it involves processing a large number of events. Especially, using point cloud processing for line tracking will result in slow performance. As the processing frequency increases, it requires solving more state variables, and the current version has not yet been parallelized. The embedding of the method into a more suitable computing architecture is part of our ongoing work.

\section*{Acknowledgments}
The authors would like to thank the fund support from the National Natural Science Foundation of China (62250610225) and Natural Science Foundation of Shanghai (22dz1201900, 22ZR1441300).

\bibliographystyle{IEEEtran}
\bibliography{citation} 

\begin{IEEEbiography}[{\includegraphics[width=1in,height=1.25in,clip,keepaspectratio]{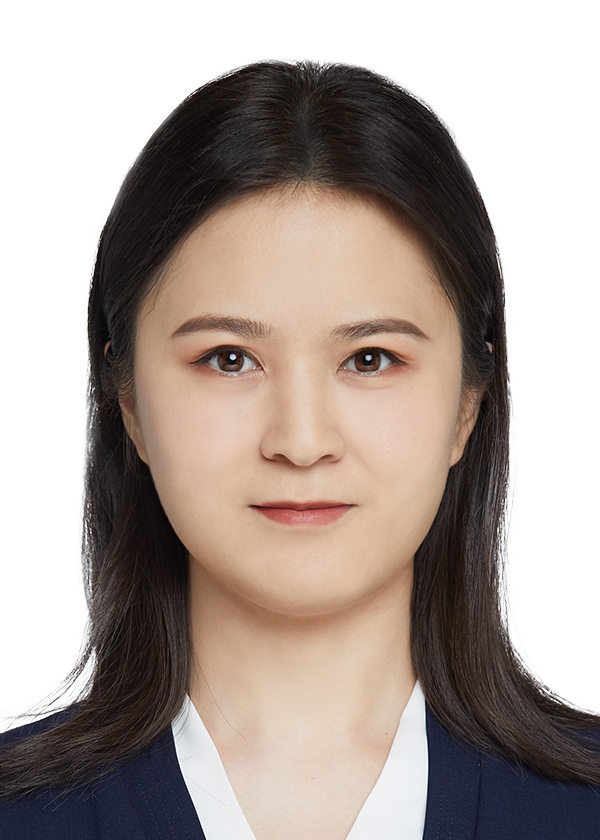}}]{Wanting Xu} Wanting Xu is currently a Ph.D. student in computer science at ShanghaiTech University, advised by Prof. Laurent Kneip. She received her B.S. degree in applied mathematics from Xinjiang University in 2018. Her research interests include visual SLAM and geometric computer vision, specifically focusing on the geometric solutions of pose estimation problems for traditional and event cameras.
\end{IEEEbiography}

\begin{IEEEbiography}[{\includegraphics[width=1in,height=1.25in,clip,keepaspectratio]{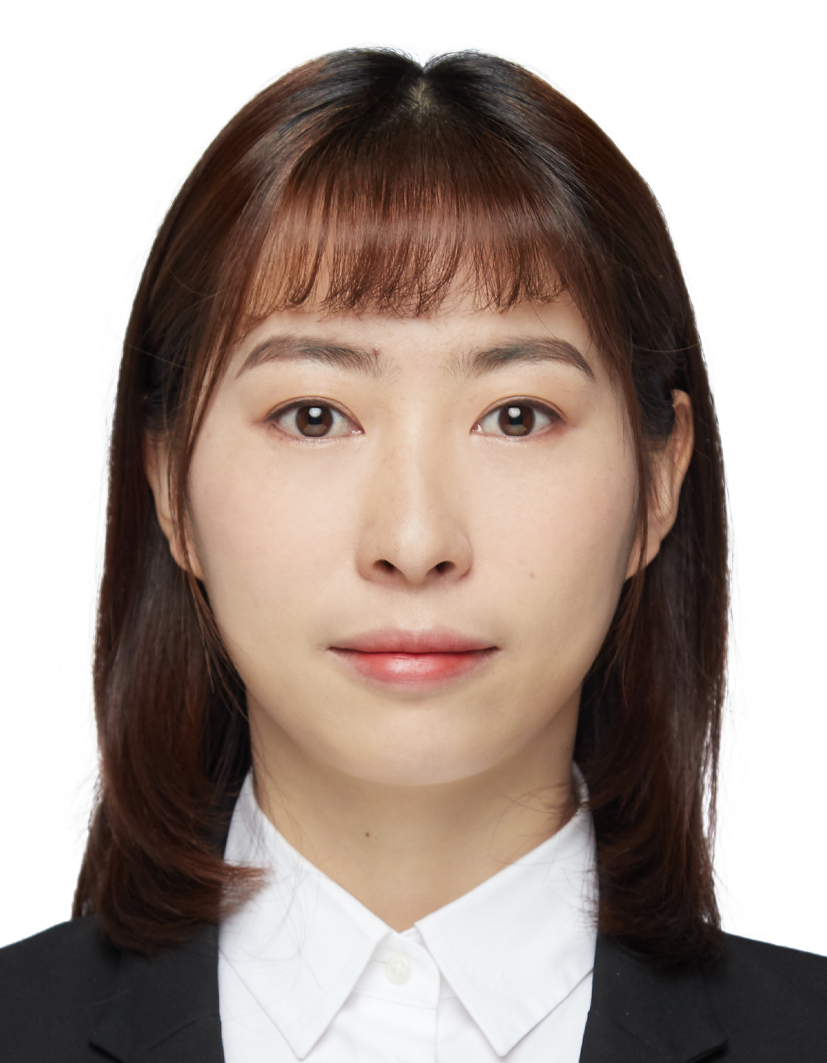}}]{Xin Peng} Xin Peng got her Ph.D. degree from the University of Chinese Academy of Sciences and ShanghaiTech University, where she worked at Mobile Perception Lab. She received a B.Eng. degree in opto-electronic technology in 2015 from the University of Electronic Science and Technology, China. Now she is working at Inc. Motovis, which is a company focusing on autonomous driving. Her research interests include event-based vision, geometric vision, pose estimation, global optimization and SLAM.
\end{IEEEbiography}

\begin{IEEEbiography}[{\includegraphics[width=1in,height=1.25in,clip,keepaspectratio]{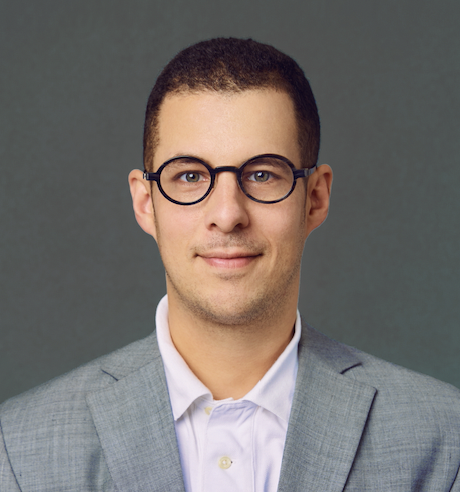}}]
{Laurent Kneip}
Laurent Kneip Dr Kneip owns a Dipl.-Ing. degree from the Friedrich-Alexander University Erlangen/N\"urnberg, and a PhD degree from ETH Zurich, where he worked at the Autonomous Systems Lab. He is also a recipient of the ARC Discovery Early Career Researcher Award (DECRA) in 2015, and the Marr Prize (honourable mention) in 2017. Dr Kneip currently is tenured Associate Professor at ShanghaiTech University, where he founded and directs the Mobile Perception Laboratory. He won the International Young Scholar Award from NSFC in 2019, the International Excellent Young Scientist Award from NSFC in 2022. He is also the director of the ShanghaiTech Automation and Robotics center. Dr Kneip has countless publications in top robotics and computer vision venues, and continuous to research on enabling intelligent mobile systems to use vision for real-time 3D perception of the environment. Dr Kneip is the main author of OpenGV.
\end{IEEEbiography}

\vfill

\end{document}